\documentclass[11pt]{article}

\usepackage[preprint]{acl}

\usepackage{booktabs}
\usepackage{array}
\usepackage{tabularx}
\usepackage{float}
\usepackage{times}
\usepackage{latexsym}
\usepackage{graphicx}
\usepackage[T1]{fontenc}
\usepackage[utf8]{inputenc}
\usepackage{microtype}
\usepackage{inconsolata}
\usepackage{booktabs}
\usepackage{amsmath}
\usepackage{pifont}
\usepackage{enumitem}
\usepackage{pgfplots}
\usepackage{pgfplotstable}
\usepackage{orcidlink}

\pgfplotsset{compat=1.18}
\usetikzlibrary{arrows.meta,calc,decorations.pathreplacing,positioning}
\renewcommand{\topfraction}{0.9}
\renewcommand{\dbltopfraction}{0.9}
\renewcommand{\textfraction}{0.07}
\renewcommand{\floatpagefraction}{0.85}
\definecolor{cFactAlign}{HTML}{2A78D6}
\definecolor{cFactGRPO}{HTML}{EDA100}
\definecolor{cFaith}{HTML}{4A3AA7}
\definecolor{cFlame}{HTML}{1BAF7A}
\definecolor{cKnowRL}{HTML}{E34948}
\definecolor{cBase}{HTML}{52514E}
\definecolor{cSupported}{HTML}{1BAF7A}
\definecolor{cNotSupported}{HTML}{303030}

\definecolor{plotlyblue}{HTML}{636EFA}
\newcommand{\cmark}{\ding{51}}
\newcommand{\xmark}{\ding{55}}

\newcommand{\flame}{\textsc{Flame}}
\newcommand{\factalign}{\textsc{FactAlign}}
\newcommand{\knowrl}{\textsc{KnowRL}}
\newcommand{\faith}{\textsc{Faith}}
\newcommand{\factgrpo}{\textsc{FactGRPO}}
\newcommand{\veriscore}{\textsc{VeriScore}}

\title{Do Factuality Methods Make LLMs More Factual? Reproducibility and Fair Comparison}

\author{
  \textbf{Pratuat Amatya\textsuperscript{1,2}\,\orcidlink{0009-0004-2593-056X}},
  \textbf{Vinay Setty\textsuperscript{1,2}\,\orcidlink{0000-0002-9777-6758}},
  \textbf{Venktesh V\textsuperscript{3}\,\orcidlink{0000-0001-5885-2175}},
  \textbf{Avishek Anand\textsuperscript{4}\,\orcidlink{0000-0002-0163-0739}}
\\
  \textsuperscript{1}Factiverse AS,
  \textsuperscript{2}University of Stavanger,
  \textsuperscript{3}Stockholm University,
  \textsuperscript{4}TU Delft
\\
  \small{
    \href{mailto:pratuat@factiverse.ai}{\texttt{pratuat@factiverse.ai}},
    \href{mailto:vinay@factiverse.ai}{\texttt{vinay@factiverse.ai}},
    \href{mailto:venktesh.viswanathan@dsv.su.se}{\texttt{venktesh.viswanathan@dsv.su.se}},
    \href{mailto:avishek.anand@tudelft.nl}{\texttt{avishek.anand@tudelft.nl}}
  }
}

\makeatletter
\providecommand\FloatBarrier{\par\begingroup\let\@elt\relax
  \edef\@tempa{\@botlist\@deferlist\@dbldeferlist}%
  \ifx\@tempa\@empty\else
    \ifx\@fltovf\relax
      \if@firstcolumn\clearpage\else\null\newpage\clearpage\fi
    \else
      \newpage\let\@fltovf\relax\FloatBarrier
    \fi
  \fi\endgroup}
\makeatother

\begin{document}
\maketitle
\begin{abstract}
Large Language Models (LLMs) are aligned to human preferences, after pre-training and supervised fine-tuning steps. However, standard RL alignment steps such as RLHF do not focus on factuality of LLMs, hence they may generate non-factual information, which could result in unintended consequences. In response, a number of factuality alignment approaches have been proposed in the literature which claim to improve factuality of LLMs on their own benchmarks. However, these methods have not been reproduced independently and are also not comparable since each uses a different base model, training data, evaluation set, judge, and metric. In this paper, we present the first independent reproduction of five factuality alignment methods, which span on-policy RL (PPO, GRPO, and Dr.~GRPO) and offline preference optimization (DPO and KTO), and compare them on a common benchmark with common evaluation metrics. We found that none of these methods reproduces in full, and on the common benchmark none of the methods significantly outperforms the base model. On closer inspection, the rewards optimize for format, length, or brevity rather than factuality and relevance, whereas the choice between on-policy and offline training matters little. Furthermore, the evaluation metrics and benchmarks are also part of the problem since they do not measure relevance of text to the prompt. Since there is no ground truth associated with the prompts, the reliance on LLM judges and retrieval for verification also makes the evaluation fragile. We release our code, data, and checkpoints.
\end{abstract}

\section{Introduction}
\pgfplotstableread{
x key published ours checkpoint
0 flame 5.60 2.60 nan
1 factalign 24.80 -2.64 2.68
2 knowrl 2.64 nan 3.71
3 factgrpo 21.20 1.10 nan
4 faith 1.20 0.33 nan
}\introgains
\pgfplotstableread{
x key bio bioLo bioHi alp alpLo alpHi sqa sqaLo sqaHi tqa tqaLo tqaHi
0 flame -1.41 -2.54 -0.26 -1.56 -3.40 0.24 -0.20 -0.80 0.40 0.73 -1.84 3.43
1 factalign -0.12 -1.06 0.82 -2.88 -4.55 -1.24 -0.20 -0.60 0.20 1.84 0.00 3.79
2 knowrl -4.46 -5.79 -3.13 -1.57 -3.95 0.67 -0.20 -1.30 0.90 -1.35 -4.41 1.71
3 factgrpo 1.38 -0.03 2.80 1.18 -1.65 4.02 1.80 0.50 3.10 -18.48 -21.79 -15.06
4 faith -33.30 -35.69 -30.87 -60.27 -63.94 -56.42 1.50 0.00 3.00 -24.60 -28.15 -21.05
}\introdeltas
\begin{figure}[t]
\centering
\begin{tikzpicture}
\begin{axis}[
  name=gains,
  width=\columnwidth, height=0.37\columnwidth,
  ybar, bar width=5pt,
  ymin=-5, ymax=27, ytick={0,10,20},
  xtick=data,
  xticklabels={\flame, \factalign, \knowrl, \factgrpo, \faith},
  xticklabel style={font=\scriptsize},
  ylabel={Gain (points)},
  ylabel style={font=\scriptsize},
  tick label style={font=\scriptsize},
  ymajorgrids=true, grid style={line width=0.2pt, draw=black!12},
  axis lines*=left, axis line style={draw=black!45}, tick style={draw=black!45},
  enlarge x limits=0.12,
  legend style={font=\scriptsize, draw=none, fill=none, at={(0.5,1.0)}, anchor=south, legend columns=3, /tikz/every even column/.append style={column sep=0.5em}},
  legend cell align=left,
  title={\scriptsize (a) Original setup, each method's own metric},
  title style={yshift=9pt},
  unbounded coords=jump,
]
\addplot[fill=black!25, draw=black!45] table[x=x, y=published] {\introgains};
\addlegendentry{published}
\addplot[fill=cFactGRPO!80, draw=cFactGRPO] table[x=x, y=ours] {\introgains};
\addlegendentry{our retrain}
\addplot[only marks, mark=o, mark size=2pt, draw=black, line width=0.7pt, forget plot] table[x expr=\thisrow{x}+0.16, y=checkpoint] {\introgains};
\addlegendimage{only marks, mark=o, mark size=2pt, draw=black, line width=0.7pt}
\addlegendentry{released checkpoint}
\draw[black!45] (axis cs:-0.5,0) -- (axis cs:4.5,0);
\end{axis}
\begin{axis}[
  name=deltas, at={(gains.south)}, anchor=north, yshift=-2.0cm,
  width=\columnwidth, height=0.41\columnwidth,
  xmin=-30, xmax=7,
  ymin=-0.6, ymax=4.6,
  ytick={0,1,2,3,4},
  yticklabels={\flame, \factalign, \knowrl, \factgrpo, \faith},
  y dir=reverse,
  xlabel={Change over the base model (points)},
  xlabel style={font=\scriptsize},
  tick label style={font=\scriptsize},
  xmajorgrids=true, grid style={line width=0.2pt, draw=black!12},
  axis lines*=left, axis line style={draw=black!45}, tick style={draw=black!45},
  legend style={font=\scriptsize, draw=none, fill=none, at={(0.5,1.0)}, anchor=south, legend columns=2, /tikz/every even column/.append style={column sep=0.5em}},
  legend cell align=left,
  title={\scriptsize (b) Common benchmark, same model, judge, and verifier},
  title style={yshift=16pt},
  error bars/x dir=both, error bars/x explicit,
  error bars/error bar style={line width=0.5pt},
  clip marker paths=true,
]
\draw[black!45] (axis cs:0,-0.6) -- (axis cs:0,4.6);
\addplot[only marks, mark=*, mark size=1.6pt, color=plotlyblue]
  table[x=bio, y expr=\thisrow{x}-0.27, x error minus expr=\thisrow{bio}-\thisrow{bioLo}, x error plus expr=\thisrow{bioHi}-\thisrow{bio}] {\introdeltas};
\addlegendentry{Biography F1@$K$}
\addplot[only marks, mark=square*, mark size=1.5pt, color=cFlame]
  table[x=alp, y expr=\thisrow{x}-0.09, x error minus expr=\thisrow{alp}-\thisrow{alpLo}, x error plus expr=\thisrow{alpHi}-\thisrow{alp}] {\introdeltas};
\addlegendentry{Alpaca-Fact F1@$K$}
\addplot[only marks, mark=triangle*, mark size=1.9pt, color=cKnowRL]
  table[x=sqa, y expr=\thisrow{x}+0.09, x error minus expr=\thisrow{sqa}-\thisrow{sqaLo}, x error plus expr=\thisrow{sqaHi}-\thisrow{sqa}] {\introdeltas};
\addlegendentry{SimpleQA acc.}
\addplot[only marks, mark=diamond*, mark size=1.9pt, color=cFaith]
  table[x=tqa, y expr=\thisrow{x}+0.27, x error minus expr=\thisrow{tqa}-\thisrow{tqaLo}, x error plus expr=\thisrow{tqaHi}-\thisrow{tqa}] {\introdeltas};
\addlegendentry{TruthfulQA acc.}
\node[font=\scriptsize, text=black!60, anchor=west] at (axis cs:-29.5,3.8) {$\leftarrow -33$ and $-60$};
\end{axis}
\end{tikzpicture}
\vspace{-2mm}
\caption{Published gains vs.\ reproduced. (a) Gain over each method's own baseline and as we measure it in the original setup, with released checkpoints as circles. (b) Change over the base model on the common benchmark with 95\% bootstrap intervals.}
\label{fig:intro}
\vspace{-3mm}
\end{figure}
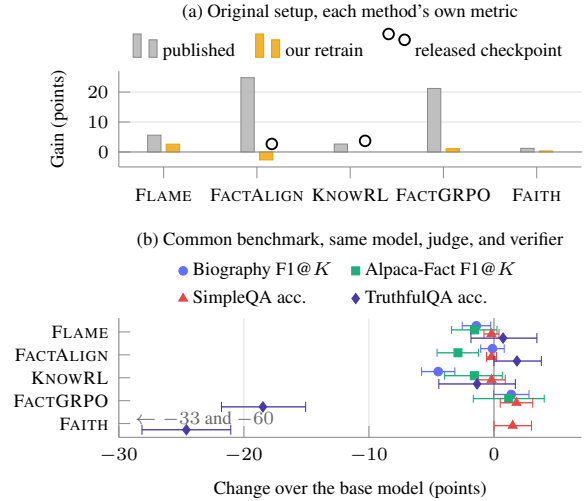

Large Language Models (LLMs) are aligned to follow instructions as part of their standard training process. But alignment methods such as Reinforcement Learning from Human Feedback (RLHF)~\citep{Christiano:2017:NeurIPS} are shown to degrade factuality of LLMs~\citep{Lin:2024:NeurIPS}. In response, a number of factuality alignment methods have emerged: factuality-aware SFT and DPO~\citep{Lin:2024:NeurIPS}, fine-grained KTO over sentence-level labels~\citep{Huang:2024:EMNLP}, reinforcement learning against a knowledge-boundary reward~\citep{Ren:2026:ACL}, a programmatic reward over long chains of thought~\citep{Chen:2026:ICML}, and PPO conditioned on an estimated knowledge state~\citep{Dong:2026:ACL}. 

However, none of these gains have been independently reproduced, and it is also not feasible to compare them with each other since these five methods use different base models, training data, evaluation sets, judges, and metrics. The improvements reported by these papers are therefore neither replicable nor generalizable.

This paper is the first reproducibility effort on all five methods. Reproducing these methods is not straightforward. For example, the authors of \flame{} do not release the code, and even when the code is released, the configuration and setup used in the code deviate from what is documented in the paper (Table~\ref{tab:availability} summarizes what the authors released). We try our best to reproduce as faithfully as possible to the setup described in the papers and document every deviation (Figure~\ref{fig:methods} and Appendix~\ref{app:reproductions}). 

To make the comparison feasible and fair, we also train every method on a single base model, Llama-3.1-8B-Instruct~\citep{Grattafiori:2024:arXiv}, with the training regime and data that its paper specifies, and evaluate on a \textbf{common benchmark} with two long-form generation tasks scored with \veriscore{}~\citep{Song:2024:EMNLP} and two short-form question answering tasks graded by an LLM judge.

Our study is driven by the following research questions:
\begin{itemize}[leftmargin=*,nosep]
\item \textbf{RQ I}: Do the published gains hold when we reproduce each method, first in its own setup and then on a common benchmark?
\item \textbf{RQ II}: How does the reward design affect long-form and short-form factuality?
\item \textbf{RQ III}: Does the training paradigm, on-policy reinforcement learning or offline preference optimization, affect factuality?
\item \textbf{RQ IV}: Do the current evaluation metrics measure factuality faithfully?
\end{itemize}

Our study yields four findings. First, \textbf{none of the methods reproduce the published results completely} when we rebuild the training pipeline independently.
For instance, the factuality gain of \flame{} reproduces, but its claim that factuality alignment costs no helpfulness does not, and \factalign{} reproduces with its released checkpoint but not when retrained.
On the common benchmark, none of the methods significantly outperforms the untrained base model.

Second, the rewards optimize for length, format, or brevity rather than factuality and relevance, and the models exploit this, which is commonly referred to as \textbf{reward hacking}.
For instance, the \factgrpo{} reward counts a claim as correct whenever evidence supports it, even if it is irrelevant to the prompt, and 36\% of its biography responses contain such claims.
The \knowrl{} and \faith{} rewards are built for short questions with a gold answer and break down on long-form prompts, where the answers of \knowrl{} grow long and imprecise and those of \faith{} shrink to one word.

Third, \textbf{the training paradigm matters less than the reward.}
On-policy RL and offline preference optimization perform similarly on the common benchmark, and the alignment step barely changes the model from the supervised fine-tuning stage.

Finally, \textbf{the evaluation itself is part of the problem.}
The factuality metrics used in benchmarks such as FActScore~\citep{Min:2023:EMNLP}, SAFE~\citep{Wei:2024:NeurIPS}, and \veriscore{}~\citep{Song:2024:EMNLP} credit claims supported by evidence irrespective of the claim's relevance to the prompt. Filtering out irrelevant claims reduces precision by 4--15 percentage points across all methods that answer at length.
Without ground truth, the scores also depend on the evidence source, the recall normalizer, and the LLM judge, each of which moves the results more than any method.

We make the code, configurations, and evaluation scripts\footnote{\url{https://anonymous.4open.science/r/factuality-alignment-reproducibility/}} as well as data and checkpoints\footnote{\url{https://huggingface.co/factualityalignment}} publicly available (anonymously).

\section{Related Work}

\subsection{Factuality Alignment}
Factuality in LLMs refers to the capability of the LLMs to generate content that adheres to factual information grounded in common sense, domain-related facts, and world knowledge \citep{survey_fctuality}.
While instruction-tuned LLMs have led to tremendous advances in a wide range of natural language tasks, they still suffer from issues such as hallucination, which might result in non-factual outputs. To improve factuality, among training-free strategies are self-consistency-based uncertainty estimation during decoding \citep{tian-etal-2023-just} and other decoding approaches \citep{decoding}. Alternatively, post-training methods such as SFT \citep{tao-etal-2024-trust} and policy optimization \citep{Dong:2026:ACL,Chen:2026:ICML,Ren:2026:ACL,Huang:2024:EMNLP} have been proposed for factuality alignment of such models. Policy optimization approaches can be on-policy or off-policy. Approaches like FactTune \citep{tian2024finetuning} construct preference pairs using generated responses and FActScore \citep{Min:2023:EMNLP} as a feedback signal. They then employ Direct Preference Optimization (DPO) in an off-policy fashion over this collected preference pairs dataset. \flame~\citep{Lin:2024:NeurIPS} also proposes a similar approach but collects self-generated traces for factuality-aware alignment, combining factuality-aware SFT and a factuality-aware RL loop through DPO. KnowTuning \citep{lyu-etal-2024-knowtuning} proposed fine-grained knowledge augmentation to identify difficult knowledge. They also employ DPO to help LLMs differentiate between reliable and unreliable knowledge.  Alternatively, on-policy methods have demonstrated that training on self-generated traces during optimization is better than training on reference traces or static self-generated traces, which may cause more hallucination due to distribution mismatch \citep{kang-etal-2025-unfamiliar,gekhman-etal-2024-fine,ghosal2024understanding}. With the success of DeepSeek employing on-policy RL using GRPO, similar RL-based tuning approaches have been proposed for factuality alignment in LLMs \citep{zhang-etal-2024-self}. While these approaches mostly focus on short-form generation, approach like \citep{Chen:2026:ICML} propose factuality-aware and aligned reasoning for long-form responses.
 \vspace{-1em}
\subsection{Factuality Evaluation}
Several evaluation setups and metrics have been proposed for factuality in short-form and long-form responses from LLMs. Evaluating the factuality of long-form responses is hard, and hence prior approaches have proposed breaking them down into atomic claims to be verified \citep{kamoi-etal-2023-wice,gao-etal-2023-rarr,guan-etal-2024-language,Wei:2024:NeurIPS}.
FActScore \citep{Min:2023:EMNLP} extracts atomic facts, which are statements that each contain a single piece of information. However, its extraction
method is specific to the domain of biographies and is inapplicable to other domains. SAFE \citep{Wei:2024:NeurIPS} extends to domains beyond biography and adopts a three-stage pipeline comprising a claim extraction stage, claim revision stage to resolve vague references, and a claim relevance check stage. While this leads to better evaluation compared to FActScore, \citet{Song:2024:EMNLP} argue that the relevance check stage also significantly removes verifiable claims, undermining evaluation of the factuality of generated text. \veriscore~\citep{Song:2024:EMNLP} was proposed to overcome these limitations and avoid the fallacy of treating all atomic claims as verifiable. \veriscore{} extracts only verifiable claims and is also one of the first methods to consider inter-sentence context when extracting claims. This also helps avoid the cost incurred by claim revision.

Existing factuality alignment approaches are not evaluated in a common setup with common benchmarks and a reliable metric. Our study primarily contributes to the rigorous evaluation of factuality alignment methods for LLMs in such a standardized setup and to an analysis of the reliability of its factuality metrics.

\section{Factual Alignment Methods}
\label{sec:methods}

No base model and no evaluation dataset is common to all five papers, and their factuality metrics differ in kind, since three are precision-like, one credits refusals, and one excludes them from its denominator (Table~\ref{tab:suite} in the appendix).
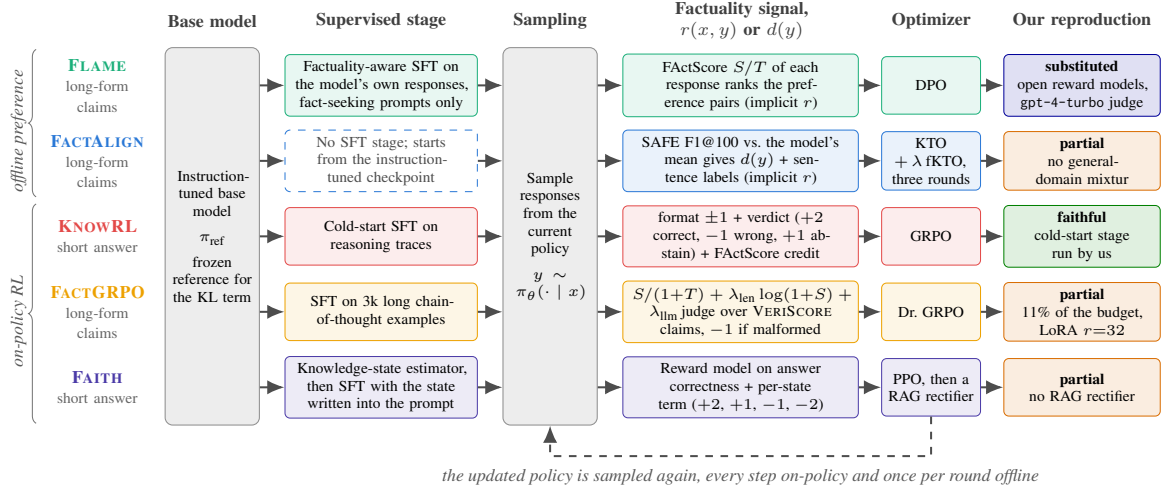
\begin{figure*}[t]
\centering
\resizebox{0.96\textwidth}{!}{%
\begin{tikzpicture}[
  x=1cm, y=1cm,
  font=\scriptsize,
  shared/.style={draw=black!45, fill=black!7, rounded corners=3pt,
    align=center, inner sep=3pt, text width=#1, font=\tiny},
  cell/.style 2 args={draw=#1, fill=#1!10, rounded corners=2pt,
    align=center, inner sep=2.5pt, text width=#2, minimum height=0.8cm,
    font=\tiny},
  nocell/.style 2 args={draw=#1, dashed, fill=white, rounded corners=2pt,
    align=center, inner sep=2.5pt, text width=#2, minimum height=0.8cm,
    text=black!70, font=\tiny},
  fid/.style 2 args={draw=#1, fill=#1!12, rounded corners=2pt,
    align=center, inner sep=2.5pt, text width=#2, minimum height=0.8cm,
    font=\tiny},
  hdr/.style={font=\scriptsize\bfseries, align=center, text width=#1,
    text=black!75},
  rowlab/.style={align=center, text width=1.3cm},
  arr/.style={-{Latex[length=2.4mm, width=2mm]}, draw=black!70,
    line width=0.8pt},
  grp/.style={font=\scriptsize\itshape, text=black!65, rotate=90,
    anchor=south, align=center},
]
\node[hdr=1.4cm] at (2.8, 0.35) {Base model};
\node[hdr=2.5cm] at (5.15, 0.35) {Supervised stage};
\node[hdr=1.4cm] at (7.5, 0.35) {Sampling};
\node[hdr=3.1cm] at (10.15, 0.35) {Factuality signal, $r(x,y)$ or $d(y)$};
\node[hdr=1.2cm] at (12.8, 0.35) {Optimizer};
\node[hdr=2.0cm] at (14.9, 0.35) {\textbf{Our reproduction}};
\node[shared=1.1cm, minimum height=5.25cm] (base) at (2.8, -2.65)
  {Instruction-tuned base model\\[3pt] $\pi_{\text{ref}}$\\[3pt]
   frozen reference for the KL term};
\node[shared=1.1cm, minimum height=5.25cm] (gen) at (7.5, -2.65)
  {Sample responses from the current policy\\[3pt]
   $y \sim \pi_\theta(\cdot \mid x)$};
\node[rowlab, text=cFlame] at (1.2, -0.55)
  {\textbf{\flame}\\{\tiny\color{black!60}long-form claims}};
\node[cell={cFlame}{2.5cm}] (s1) at (5.15, -0.55)
  {Factuality-aware SFT on the model's own responses, fact-seeking
   prompts only};
\node[cell={cFlame}{3.1cm}] (f1) at (10.15, -0.55)
  {FActScore $S/T$ of each response ranks the preference pairs (implicit $r$)};
\node[cell={cFlame}{1.2cm}] (o1) at (12.8, -0.55) {DPO};
\node[rowlab, text=cFactAlign] at (1.2, -1.6)
  {\textbf{\factalign}\\{\tiny\color{black!60}long-form claims}};
\node[nocell={cFactAlign}{2.5cm}] (s2) at (5.15, -1.6)
  {No SFT stage; starts from the instruction-tuned checkpoint};
\node[cell={cFactAlign}{3.1cm}] (f2) at (10.15, -1.6)
  {SAFE F1@100 vs.\ the model's mean gives $d(y)$ + sentence labels (implicit $r$)};
\node[cell={cFactAlign}{1.2cm}] (o2) at (12.8, -1.6)
  {KTO $+\,\lambda\,$fKTO, three rounds};
\node[rowlab, text=cKnowRL] at (1.2, -2.65)
  {\textbf{\knowrl}\\{\tiny\color{black!60}short answer}};
\node[cell={cKnowRL}{2.5cm}] (s3) at (5.15, -2.65)
  {Cold-start SFT on reasoning traces};
\node[cell={cKnowRL}{3.1cm}] (f3) at (10.15, -2.65)
  {format $\pm1$ + verdict ($+2$ correct, $-1$ wrong, $+1$ abstain) + FActScore credit};
\node[cell={cKnowRL}{1.2cm}] (o3) at (12.8, -2.65) {GRPO};
\node[rowlab, text=cFactGRPO] at (1.2, -3.7)
  {\textbf{\factgrpo}\\{\tiny\color{black!60}long-form claims}};
\node[cell={cFactGRPO}{2.5cm}] (s4) at (5.15, -3.7)
  {SFT on 3k long chain-of-thought examples};
\node[cell={cFactGRPO}{3.1cm}] (f4) at (10.15, -3.7)
  {$S/(1{+}T) + \lambda_{\text{len}}\log(1{+}S) + \lambda_{\text{llm}}\,\text{judge}$ over \veriscore{} claims, $-1$ if malformed};
\node[cell={cFactGRPO}{1.2cm}] (o4) at (12.8, -3.7) {Dr.\ GRPO};
\node[rowlab, text=cFaith] at (1.2, -4.75)
  {\textbf{\faith}\\{\tiny\color{black!60}short answer}};
\node[cell={cFaith}{2.5cm}] (s5) at (5.15, -4.75)
  {Knowledge-state estimator, then SFT with the state written into the
   prompt};
\node[cell={cFaith}{3.1cm}] (f5) at (10.15, -4.75)
  {Reward model on answer correctness + per-state term ($+2$, $+1$, $-1$, $-2$)};
\node[cell={cFaith}{1.2cm}] (o5) at (12.8, -4.75)
  {PPO, then a RAG rectifier};
\node[fid={blue!60!black}{2.0cm}] (r1) at (14.9, -0.55)
  {\textbf{substituted}\\ open reward models, \texttt{gpt-4-turbo} judge};
\node[fid={orange!85!black}{2.0cm}] (r2) at (14.9, -1.6)
  {\textbf{partial}\\ no general-domain mixtur};
\node[fid={green!50!black}{2.0cm}] (r3) at (14.9, -2.65)
  {\textbf{faithful}\\ cold-start stage run by us};
\node[fid={orange!85!black}{2.0cm}] (r4) at (14.9, -3.7)
  {\textbf{partial}\\ 11\% of the budget, LoRA $r{=}32$};
\node[fid={orange!85!black}{2.0cm}] (r5) at (14.9, -4.75)
  {\textbf{partial}\\ no RAG rectifier};
\foreach \i in {1,...,5} {
  \draw[arr] (base.east |- s\i.west) -- (s\i.west);
  \draw[arr] (s\i.east) -- (gen.west |- s\i.east);
  \draw[arr] (gen.east |- f\i.west) -- (f\i.west);
  \draw[arr] (f\i.east) -- (o\i.west);
  \draw[arr] (o\i.east) -- (r\i.west);
}
\draw[arr, dashed] (o5.south) -- ++(0, -0.55) -| (gen.south);
\node[fill=white, inner sep=2pt, font=\scriptsize\itshape, text=black!65]
  at (10.15, -6.0)
  {the updated policy is sampled again, every step on-policy and once per
   round offline};
\draw[decorate, decoration={brace, amplitude=3pt}, black!55]
  (0.35, -2.1) -- (0.35, -0.05)
  node[grp, midway, xshift=-3pt] {offline preference};
\draw[decorate, decoration={brace, amplitude=3pt}, black!55]
  (0.35, -5.25) -- (0.35, -2.2)
  node[grp, midway, xshift=-3pt] {on-policy RL};
\end{tikzpicture}}
\vspace{-2mm}
\caption{The five alignment methods and our reproductions. Gray boxes show shared stages, and colored boxes show differences between methods. \flame{} and \factalign{} use offline preference optimization; the other three methods use on-policy RL with explicit rewards. The final column records the main deviation in each reproduction. Further details appear in Section~\ref{sec:original} and Appendix~\ref{app:reproductions}.}
\label{fig:methods}
\vspace{-3mm}
\end{figure*}

\subsection{Method Descriptions}

Let  $\pi_\theta$ denote the LLM-based policy being aligned and $\pi_{\text{ref}}$
for its frozen initialization. A prompt / question $x$ elicits a response $y$. For
long-form evaluation $y$ decomposes into atomic claims
$c_1,\dots,c_T$, of which $S$ are judged supported against an evidence
source; precision is $P = S/T$, recall at a fixed budget $K$ is
$R_K = \min(1, S/K)$, and $\text{F1}@K$ is their harmonic mean. We
follow \citet{Chen:2026:ICML} in also calling $S$ the \emph{detail
level}, which indicates the number of correct facts in the response. For short-form
evaluation a response is graded \textsc{correct}, \textsc{incorrect}, or
\textsc{not attempted}. We write $r(x,y)$ for a scalar
reward and $d(y) \in \{+,-\}$ for a binary desirability label.

Figure~\ref{fig:methods} summarizes the five pipelines, which differ in the signal of the reward $r$, the granularity at which it is applied, and what is held fixed while it is optimized.

\paragraph{\flame~\citep{Lin:2024:NeurIPS}} makes both
alignment stages factuality-aware. In the SFT stage, each instruction from seed data is classified as being fact-based or not. For fact-based instructions $x$, the SFT targets are the
policy's \emph{own} generations instead of external text, to prevent hallucinations from training on unknown data. For non-fact-based instructions, the authors propose to leverage human responses. For DPO~\citep{Rafailov:2023:NeurIPS} training, responses to fact-based instructions are ranked by factual precision, $r_{\text{fact}} = S/T$. Responses to other instructions are ranked by an instruction-following reward. The preference data are generated before training and remain fixed as the policy is updated. Training is therefore off-policy, and the responses in the preference data may become less representative of the current policy.

\noindent \textbf{\factalign~\citep{Huang:2024:EMNLP}} replaces
pairwise preference with KTO (Kahneman-Tversky Optimization)~\citep{Ethayarajh:2024:ICML}, which utilizes binary labels,
$d(y)$ instead of pairwise preferences. Since the responses with corresponding rewards are collected offline which may cause discrepancy between assessed responses and the model being trained, the authors also propose an iterative optimization mechanism. Here, new samples are periodically sampled from the model being trained.   A response is considered as desirable when its SAFE
$\text{F1}@100$ exceeds a threshold set at the model's own mean. Apart from response-level reward, the work's
contribution is also fine-grained sentence-level factuality alignment through \emph{fine-grained} KTO (fKTO). Hence the loss also comprises a sentence-level term,
$\mathcal{L} = \mathcal{L}_{\text{KTO}} +
\lambda\,\mathcal{L}_{\text{fKTO}}$, with separate KL strengths $\beta$
and $\beta_{\text{fg}}$.\looseness=-1 

\noindent \textbf{\knowrl~\citep{Ren:2026:ACL}} adopts GRPO
\citep{Shao:2024:arXiv} on top of a cold-start SFT stage, with factual supervision introduced into the chain-of-thought (CoT) reasoning process. The reward is of the form
$r = r_{\text{format}} + r_{\text{correct}} + r_{\text{fact}}$. The correctness term gives $+2$ for a correct answer, $-1$ for an incorrect answer, and $+1$ for an explicit non-attempt. Abstaining therefore receives a higher reward than answering incorrectly. The authors argue that the cold-start stage is needed to keep training stable under this reward.\looseness=-1

\noindent \textbf{\factgrpo~\citep{Chen:2026:ICML}} optimizes a \emph{programmatic}
reward with Dr.\ GRPO~\citep{Liu:2025:arXiv} over a long chain of thought,
\begin{equation*}
r = \frac{S}{1+T} + \lambda_{\text{len}}\log(1+S)
    + \lambda_{\text{llm}}\,\text{judge},
\end{equation*}

with $r = -1$ for a malformed chain of thought. The first term measures factual precision ($\mathcal{R}_{fact}$), with $1$ added to the denominator to avoid division by zero when no claims are extracted. The second term rewards detail ($\mathcal{R}_{dtl}$), measured by the number of supported claims $S$ in the response, through $\log(1+S)$. The third term measures the response's relevance to the question ($\mathcal{R}_{rel}$) using an LLM judge. The weights $\lambda_{\text{len}}$ and $\lambda_{\text{llm}}$ control the contributions of detail and relevance to the total reward. For $T>0$, we can write the first term as
$S/(1+T)=PT/(1+T)$, where $P=S/T$. At a fixed precision, this term increases with the number of claims. Thus, even with $\lambda_{\text{len}}=0$, as in the released configuration, the reward retains a preference for more detailed responses of the same precision and does not remove  incentives to produce more claims.\looseness=-1

\noindent \textbf{\faith~\citep{Dong:2026:ACL}} conditions generation on an estimated \emph{knowledge state}. For each training question, the method samples multiple responses and computes their consistency and semantic entropy. These statistics determine one of four states, defined by whether the model is estimated to possess the knowledge ($K$ or $\neg K$) and whether its answer reflects that knowledge ($H$ or $\neg H$). The paper calls these dimensions ``trustworthiness'' and ``honestness.'' The estimated state is written into the prompt, and PPO~\citep{Schulman:2017:arXiv} uses an additional state-dependent reward of $+2$, $+1$, $-1$, or $-2$.\looseness=-1

The methods differ in both the unit of supervision and their treatment of abstention. \flame{}, \factalign{}, and \factgrpo{} score claims extracted from long-form responses. \knowrl{} and \faith{} score short answers and give credit for abstention. \knowrl{} assigns $+1$ to a non-attempt and $-1$ to an incorrect answer, while \faith{} rewards appropriate refusals. In \factgrpo{}, $S/(1+T)$ rewards more claims at fixed positive precision.

\section{Reproduction in the Original Setups}
\label{sec:original}
\label{sec:origsetup}

To answer \textbf{RQ I}, we first reproduce each method in its original setup, with the released code, data, and checkpoints (where available), and record for every component whether it is faithful, substituted, or partial (last column of Figure~\ref{fig:methods}; Appendix~\ref{app:reproductions} lists every deviation). Tables~\ref{tab:pipelines} and~\ref{tab:suite} in the appendix list the training stages, data, and metrics of each paper. We contacted the authors of all five papers about the discrepancies between the released artifacts and the papers, and only the \faith{} authors had replied at the time of writing (Appendix~\ref{app:faith}). Table~\ref{tab:papervsours} sets the published numbers against ours. We observe that all methods except \faith{} and \factgrpo{} follow a similar trend as their respective original implementations, although none reproduces their published results in full.\looseness=-1

\noindent \textbf{\flame{}.}
We retrain all four models at 70B with open reward models and \texttt{gpt-4-turbo} in place of the unreleased reward models and the retired judge. The factuality claim reproduces, with a biography FActScore above the published one, whereas the claim that factuality alignment costs no helpfulness does not, since our AlpacaEval win rate falls well below the published one (Appendix~\ref{app:flame}).

\noindent \textbf{\factalign{}.}
The released checkpoint improves over its base under two retrievers, as published. Our retrain on gemma-2b for three iterations without the general-domain mixture lands at the level of the paper's own ablation without it, so the gain depends on that mixture rather than on the objective alone (Appendix~\ref{app:factalign}).\looseness=-1

\noindent \textbf{\knowrl{}.}
Both our run with the cold-start stage followed by 150 GRPO steps and our evaluation of the released checkpoint reproduce the reported behavior without refusal collapse, albeit with lower converged rewards. In contrast, all runs using the released learning rate or omitting the cold-start stage collapse into refusals. However, closer inspection reveals that the released checkpoint abstains on most SimpleQA questions, and its score increases mechanically as refusals are excluded from the metric (Appendix~\ref{app:knowrl}).

\paragraph{\factgrpo{}.}
With 400 steps of Dr.\ GRPO, a tenth of the published optimization, and a rank-32 adapter, precision rises slightly while the supported count falls, against the published gains in both, and the reward stays flat, so this run bounds a tenth of the budget rather than the method (Appendix~\ref{app:factgrpo}). The rerun with the relevance term of the reward wired, which the first run had left out, behaves the same and is the variant used in Section~\ref{sec:commonresults}.

\paragraph{\faith{}.}
With four of the five stages on Llama-3-8B, without the rectifier and with the estimator-predicted state, the PPO gains reproduce on two of four datasets, at a fraction of the published size on average, and the model never refuses. Its SFT targets are the short gold answers of question answering sets and its reward model accepts a prediction that the gold answer contains, so both stages pull the policy toward one-word answers, which is what we measure on the long-form benchmarks (Appendix~\ref{app:faith}).\looseness=-1

\begin{table}[t]
\centering
\small
\footnotesize
\setlength{\tabcolsep}{3pt}
\renewcommand{\arraystretch}{0.92}
\begin{tabular}{@{}p{0.50\columnwidth} r r c@{}}
\toprule
\textbf{Quantity} & \textbf{Paper} & \textbf{Ours} & \\
\midrule
\multicolumn{4}{@{}l}{\flame} \\
  Bio FActScore, $\textsc{Sft}^{\text{f}}{+}\textsc{Dpo}^{\text{f}}$ &
    47.9 & 54.2 & \cmark \\
  AlpacaEval win rate vs.\ SFT${+}$DPO & 51.2 & 37.7 & \xmark \\
\multicolumn{4}{@{}l}{\factalign, $\text{F1}@100$} \\
  released checkpoint vs.\ base, paired &
    --- & $+2.68$ & \cmark \\
  our three-iteration run vs.\ its baseline &
    $+24.8$ & $-2.64$ & \xmark \\
  \quad paper's ``$-$ general-domain'' ablation &
    $-0.41$ & $-2.64$ & \cmark \\
\multicolumn{4}{@{}l}{\knowrl} \\
  refusal rate, final fifth of training & no collapse & 0.2\% & \cmark \\
  combined reward at convergence & ${\approx}{-}0.1$ & $-0.637$ & $\approx$ \\
  released checkpoint, SimpleQA & 1.92$\to$4.56 & 2.00$\to$5.71 & \cmark \\
\multicolumn{4}{@{}l}{\factgrpo~\citep{Chen:2026:ICML}, precision / detail level} \\
  Biography precision & 25.9$\to$47.1 & 35.8$\to$36.9 & \xmark \\
  Alpaca-Fact precision & 60.2$\to$79.7 & 61.9$\to$64.7 & \xmark \\
  Biography detail level & 10.3$\to$14.8 & 12.1$\to$10.9 & \xmark \\
  Alpaca-Fact detail level & 30.7$\to$36.9 & 20.0$\to$18.1 & \xmark \\
\multicolumn{4}{@{}l}{\faith, precision, \faith\textsubscript{sft}$\to$\faith\textsubscript{sft+ppo}} \\
  mean gain over four datasets (two reproduce) & $+1.20$ & $+0.33$ & $\approx$ \\
  refusal rate & --- & $0.000$ & \xmark \\
\bottomrule
\end{tabular}
\vspace{-1mm}
\caption{Published and reproduced results under each method's protocol, in points unless noted. Compare within rows only. \cmark: reproduced; \xmark: not reproduced; $\approx$: same direction, different magnitude.}
\label{tab:papervsours}
\vspace{-3mm}
\end{table}

\section{Common Benchmark}
\label{sec:commonresults}

\subsection{Setup}
\label{sec:setup}

\paragraph{Model and fine-tuning} All five methods of Section~\ref{sec:methods} are trained on Llama-3.1-8B-Instruct, each alignment stage as a LoRA~\citep{Hu:2022:ICLR} adapter of rank 32 on top of the method's own supervised stage with the hyperparameters of its paper. Since the five methods carry too many settings to print, the full configuration of every run is listed in the code repository and summarized in Appendix~\ref{app:repro}. All methods generate with greedy decoding from the bare question or instruction, with at most 1,024 new tokens for long-form prompts and 256 for short-form questions, and \faith{} additionally receives its estimator-predicted knowledge state in the prompt, as its paper prescribes.
\noindent \textbf{Benchmarks} Short-form evaluation uses the first 1,000 SimpleQA questions~\citep{Wei:2024:arXiv} and all 817 TruthfulQA questions~\citep{Lin:2022:ACL}, graded as correct, incorrect, or not attempted by \texttt{gpt-4o} using the official SimpleQA prompt. We report accuracy, abstention rate, accuracy on attempted questions, and their harmonic mean $F$. Long-form evaluation uses 495 \textsc{FActScore} biography prompts~\citep{Min:2023:EMNLP} and 253 Alpaca-Fact instructions, selected from AlpacaEval by the \flame{} fact-seeking classifier following \citet{Lin:2024:NeurIPS}. We score responses with \veriscore{}~\citep{Song:2024:EMNLP}, which extracts verifiable claims and checks each against ten web search results. We report the average numbers of supported and unsupported claims per response, precision (assigning zero to responses with no claims), and $\text{F1}@K$, with $K$ fixed at 20 and 19, the median claim counts of an earlier pool (Section~\ref{sec:rq4}). For each long-form table, we evaluate all models on the same subset of prompts; excluded prompts are listed in Appendix~\ref{app:repro}. We test significance against the base model using McNemar's exact test for short-form evaluation and paired bootstrap resampling for long-form precision, recall, and $\text{F1}@K$, with 10{,}000 samples. We apply Holm's correction for multiple comparisons.

\subsection{Overall Results}


\begin{table}[t]
\centering
\footnotesize
\setlength{\tabcolsep}{4pt}
\renewcommand{\arraystretch}{0.92}
\begin{tabular}{@{}l cccc@{}}
\toprule
Method & Acc. & Abst. & Acc$\mid$att. & F \\
\midrule
\multicolumn{5}{@{}l}{SimpleQA ($n = 1{,}000$)} \\
base (untrained) & .034 & .812 & .181 & .057 \\
\flame           & .032 & \textbf{.851} & \textbf{.215} & .056 \\
\factalign       & .032 & .813 & .171 & .054 \\
\knowrl          & .032 & .822 & .180 & .054 \\
\factgrpo        & \textbf{.052} & .414 & .089 & \textbf{.066} \\
\faith           & .049 & .000 & .049 & .049 \\
\addlinespace
\multicolumn{5}{@{}l}{TruthfulQA ($n = 817$)} \\
base (untrained) & .416 & .108 & .466 & .440 \\
\flame           & .424 & .108 & .475 & .448 \\
\factalign       & \textbf{.435} & .111 & \textbf{.489} & \textbf{.460} \\
\knowrl          & .403 & .124 & .459 & .429 \\
\factgrpo        & .231 & \textbf{.368} & .366 & .284 \\
\faith           & .170 & .029 & .175 & .173 \\
\bottomrule
\end{tabular}

\vspace{-1mm}
\caption{Short-form evaluation, one judge (\texttt{gpt-4o}) across every method. \emph{Abst.}\ is the share of questions the model
declined; \emph{Acc$\mid$att.}\ is accuracy over attempted questions
only, which is the quantity a method can raise by declining more.}
\label{tab:shorttrack}
\vspace{-3mm}
\end{table}

\begin{table*}[t]
\centering
\footnotesize
\setlength{\tabcolsep}{4pt}
\renewcommand{\arraystretch}{0.92}
\begin{tabular}{lrrrrrrrrr}
\toprule
 & & \multicolumn{4}{c}{\veriscore} & \multicolumn{4}{c}{\veriscore{}+ Relevance} \\
\cmidrule(lr){3-6} \cmidrule(lr){7-10}
Method & \#Claims & \#Sup. & P & R & F1 & \#Sup. & P & R & F1 \\
\midrule
\multicolumn{10}{l}{\textit{Biography ($n = 490$, $K = 20$)}} \\
base (untrained) & 27.6 & \textbf{12.0} & 0.351 & \textbf{0.472} & 0.370 & \textbf{10.6} & 0.283 & \textbf{0.434} & 0.325 \\
\midrule
\flame & 23.5 & 10.6 & 0.338 & 0.436\rlap{$^{\dagger}$} & 0.356\rlap{$^{\dagger}$} & 9.7 & 0.294 & 0.407\rlap{$^{\dagger}$} & 0.324 \\
\factalign & 27.1 & 11.7 & 0.343 & 0.468 & 0.368 & 10.4 & 0.284 & 0.431 & 0.326 \\
\knowrl & \textbf{29.8} & 11.3 & 0.281\rlap{$^{\dagger}$} & 0.440\rlap{$^{\dagger}$} & 0.325\rlap{$^{\dagger}$} & 10.0 & 0.237\rlap{$^{\dagger}$} & 0.409\rlap{$^{\dagger}$} & 0.291\rlap{$^{\dagger}$} \\
\factgrpo & 25.4 & 10.1 & 0.342 & 0.466 & \textbf{0.383} & 9.0 & 0.301 & 0.423 & \textbf{0.342}\rlap{$^{\dagger}$} \\
\faith & 0.6 & 0.4 & \textbf{0.374} & 0.019\rlap{$^{\dagger}$} & 0.037\rlap{$^{\dagger}$} & 0.4 & \textbf{0.372}\rlap{$^{\dagger}$} & 0.019\rlap{$^{\dagger}$} & 0.036\rlap{$^{\dagger}$} \\
\midrule
\multicolumn{10}{l}{\textit{Alpaca-Fact ($n = 248$, $K = 19$)}} \\
base (untrained) & 29.4 & 19.9 & 0.618 & 0.681 & 0.615 & 15.9 & 0.491 & 0.586 & \textbf{0.500} \\
\midrule
\flame & 26.6 & 18.2 & 0.621 & 0.653\rlap{$^{\dagger}$} & 0.599 & 14.5 & \textbf{0.500} & 0.565 & 0.492 \\
\factalign & 28.1 & 18.1 & 0.598 & 0.651\rlap{$^{\dagger}$} & 0.586\rlap{$^{\dagger}$} & 14.4 & 0.484 & 0.556\rlap{$^{\dagger}$} & 0.478 \\
\knowrl & \textbf{33.1} & \textbf{20.8} & 0.582 & \textbf{0.689} & 0.599 & \textbf{16.9} & 0.471 & \textbf{0.612} & \textbf{0.500} \\
\factgrpo & 25.4 & 17.6 & \textbf{0.624} & 0.677 & \textbf{0.626} & 13.2 & 0.471 & 0.566 & 0.489 \\
\faith & 0.1 & 0.1 & 0.101\rlap{$^{\dagger}$} & 0.006\rlap{$^{\dagger}$} & 0.012\rlap{$^{\dagger}$} & 0.1 & 0.097\rlap{$^{\dagger}$} & 0.006\rlap{$^{\dagger}$} & 0.012\rlap{$^{\dagger}$} \\
\bottomrule
\end{tabular}

\vspace{-1mm}
\caption{Pooled results for long-form factuality under \veriscore{} and \veriscore{} with relevance filtering evaluation. \#Claims is the mean number of extracted claims per response and \#Sup.\ the mean number of supported claims; bold marks the highest value per column within each dataset. † marks a significant difference from base (paired bootstrap test, Holm-corrected $p < 0.05$).}
\label{tab:factuality}
\vspace{-4mm}
\end{table*}

The common evaluation setup employs one base model, one decoding configuration, one judge, and one claim verifier, across all methods so that differences are attributable to training rather than to evaluation. We found that no method improves significantly over the untrained model on any of the four datasets, apart from two long-form cases discussed below. From Table~\ref{tab:shorttrack}, we observe that \faith{} and \factgrpo{} lose about 25 and 18 points of TruthfulQA accuracy, while the other three methods stay within two points of the base model, which is not statistically significant.\looseness=-1

On long-form evaluation, \faith{} has the highest precision on the biography task, but the gain is an artifact of 0.4 supported claims in a one-sentence (very brief) response, which is why its relevance-filtered precision is significantly above the base model, but it has the lowest $\text{F1}@K$ (Table~\ref{tab:factuality}). Since \faith{} is proposed as a general factuality alignment method for LLMs and not as specific to short-form tasks, the long-form benchmarks are within its stated scope. \factgrpo{} has the highest $\text{F1}@K$ on both datasets with fewer claims than the base model. Its relevance-filtered $\text{F1}@K$ on biographies is .017 above the base model and is the only gain apart from the \faith{} artifact that is statistically significant, which is a small fraction of the published improvement. On SimpleQA, \flame{} raises its accuracy over attempted questions by declining more often while its overall accuracy stays unchanged, so the accuracy alone understates how differently these models behave. Hence, none of the methods provides substantial factuality gains in our common evaluation setting.\looseness=-1

\subsection{Reward Design}
\label{sec:reward}

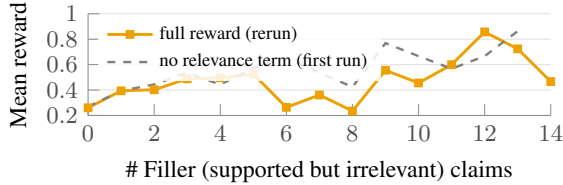
\begin{figure}[t]
\centering
\begin{tikzpicture}
\begin{axis}[
  width=\columnwidth, height=0.38\columnwidth,
  xlabel={\# Filler (supported but irrelevant) claims},
  ylabel={Mean reward},
  xmin=0, xmax=14, ymin=0.2, ymax=1.0,
  xtick={0,2,...,14}, ytick={0.2,0.4,0.6,0.8,1.0},
  axis lines*=left,
  axis line style={draw=black!45},
  tick style={draw=black!45},
  tick label style={font=\footnotesize, text=black!70},
  label style={font=\footnotesize},
  ymajorgrids=true,
  grid style={line width=0.2pt, draw=black!12},
  y tick label style={/pgf/number format/fixed, /pgf/number format/precision=1},
  legend style={font=\scriptsize, draw=none, fill=white, fill opacity=0.85, text opacity=1, at={(0.02,0.99)}, anchor=north west},
  legend cell align=left,
]
\addplot[color=cFactGRPO, line width=1.1pt, mark=square*, mark size=1.1pt] coordinates {
    (0,0.2603) (1,0.3939) (2,0.4024) (3,0.4883) (4,0.4960) (5,0.5255)
    (6,0.2637) (7,0.3614) (8,0.2353) (9,0.5556) (10,0.4555) (11,0.6000)
    (12,0.8571) (13,0.7241) (14,0.4667)
};
\addlegendentry{full reward (rerun)}
\addplot[color=black!50, dashed, line width=0.9pt, mark=none] coordinates {
    (0,0.2688) (1,0.3945) (2,0.4439) (3,0.5361) (4,0.4430) (5,0.5726)
    (6,0.6554) (7,0.5375) (8,0.4241) (9,0.7692) (10,0.6667) (11,0.5676)
    (12,0.6667) (13,0.8631)
};
\addlegendentry{no relevance term (first run)}
\end{axis}
\end{tikzpicture}
\vspace{-2mm}
\caption{Mean \factgrpo{} reward as a function of the number of filler (supported but irrelevant) claims in a long-form biography response, for the rerun with the full reward and for the first run without the relevance term. Bins beyond 14 hold one or two responses and are not shown.}
\label{fig:factgrpo_reward_hacking}
\vspace{-6mm}
\end{figure}

To answer \textbf{RQ II}, we analyze the effect of rewards on responses. The \factgrpo{} reward is a function of the number of supported and total claims, with a relevance term that contributes little, so a claim that is true but does not address the prompt earns the same reward as one that does. We call such claims filler claims. They occur in about two fifths of the long-form responses of every method that answers at length (Figure~\ref{fig:filler_claim_distribution} in the appendix), and the \factgrpo{} reward rises with their number (Figure~\ref{fig:factgrpo_reward_hacking}), since the factuality term $\mathcal{R}_{fact}$ dominates the relevance term $\mathcal{R}_{rel}$ (Table~\ref{tab:rfact-dominates-rrel} in the appendix). The model can therefore hack the reward by generating claims that are factual but off-topic. The \knowrl{} reward adds a format term, a verdict term, and a factuality term, but on free-form prompts only the verdict term carries a signal, since the model never emits the expected tags and the factuality term retrieves Wikipedia pages by title, which questions do not have. The \faith{} reward credits brevity, as described above.

When we score every method's generations with every method's reward, normalized by response length, the rewards disagree with each other and with the benchmarks (Table~\ref{tab:crossreward} in the appendix). Only the \factalign{} and \factgrpo{} rewards prefer their own generations, the latter only because every other method fails its format check, whereas the \flame{} reward ranks the base model's generations above \flame{}'s own. Where the \knowrl{} and \faith{} rewards prefer a method over the base model, it is another method's generations, as both rate \flame{}'s generations above the base model's, which none of the benchmarks does. The rewards therefore credit length, format, or brevity rather than the factuality that the benchmarks measure.\looseness=-1

\subsection{Training Paradigm}
\label{sec:sftdecomp}
To answer \textbf{RQ III}, we group the methods by paradigm. \knowrl{}, \factgrpo{}, and \faith{} optimize on-policy with GRPO, Dr.\ GRPO, and PPO, whereas \flame{} and \factalign{} optimize offline with DPO and KTO on preference data. Neither group separates from the base model on the common benchmark, and the two large losses on TruthfulQA both come from on-policy methods whose training changed the output format. To separate the paradigm from the supervised data it starts from, Table~\ref{tab:sftonly} shows each supervised stage on its own and Table~\ref{tab:shorttrack} the pipeline that follows it. On the short-form benchmarks the alignment step adds almost nothing, since no method moves by more than .007 in accuracy and none of the differences is significant, although individual answers do change, as \factgrpo{} turns 38 TruthfulQA items from incorrect to correct and 42 the other way. On the long-form benchmarks (Table~\ref{tab:longdecomp} against Table~\ref{tab:factuality}) the alignment step moves precision and $\text{F1}@K$ by less than three points, and none of the differences is significant. We believe this is because the supervised stage already sets the format and the length of the answers, and the objective that follows has little room to change what the metrics count.\looseness=-1

\section{Reliability of Factuality Evaluation}
\label{sec:rq4}

To answer \textbf{RQ IV}, we examine short- and long-form evaluation pipelines whose components are stochastic, lack ground truth, or depend on the model pool. We thus treat RQ IV as a measurement question rather than a flaw of any single metric.

\paragraph{Relevance.}
\veriscore{} credits supported claims even when they are irrelevant or repeated (Appendix, Table~\ref{tab:eval-limitations}). Relevance filtering lowers precision by 4--7 points on biographies and 11--15 on Alpaca-Fact for methods producing long responses (Table~\ref{tab:factuality}). Filler claims occur in 178--223 of 490 biography responses per method, but only once for \faith{} (Appendix Figures~\ref{fig:filler_claim_distribution} and~\ref{fig:hallucinated_claim_distribution}). The \factgrpo{} reward also credits these claims, enabling the reward hacking discussed under RQ II (Figure~\ref{fig:factgrpo_reward_hacking}).\looseness=-1

\paragraph{Extraction and scoring conventions.}
Two scoring runs return identical claim lists for fewer than half of the same responses. We use paired tests, though these do not remove extraction noise. Zero-claim responses receive zero precision; excluding them raises \faith{}'s Alpaca-Fact precision from 0.101 to 0.781.\looseness=-1

\paragraph{Recall normalizer.}
Changing the model pool changes its median claim count $K$, shifting $\text{F1}@K$ by more than .1 for identical responses (Table~\ref{tab:ksens}). We fix $K$ at 20 for biographies and 19 for Alpaca-Fact. Appendix~\ref{app:ksens} examines the medians after adding the \factgrpo{} rerun.\looseness=-1

\paragraph{Evidence source.}
A true claim without retrieved support receives the same label as a false one. Re-verifying biography claims against the subject's Wikipedia article lowers precision by about twenty points for methods producing long responses and changes their ranking (Appendix~\ref{app:reference}). The article passages support only two fifths of the claims supported by web search. Changing the evidence source thus affects scores more than changing the method.\looseness=-1

\paragraph{Judge.}
The short-form judge decides the absolute numbers, since regrading with \texttt{gpt-oss-120b} changes up to a third of the TruthfulQA labels (Table~\ref{tab:openjudge}), although the large drops of \factgrpo{} and \faith{} persist.\looseness=-1

\paragraph{Human audit.}
Two of the authors labeled the same 300 claims independently, 50 per variant and balanced over the automatic labels, without seeing those labels (Appendix~\ref{app:audit}), and both judged 24 of them not to be claims at all.
\veriscore{} agrees with their consensus on 84\% of the claims (Cohen's $\kappa = 0.65$), counting claims that the evidence does not settle as unsupported, while the annotators agree with each other on 78\% ($\kappa = 0.50$).
Where both annotators agree, \veriscore{} is right for 97\% of the claims it accepts but only 65\% of those it rejects, so its precision is, if anything, an underestimate.
The relevance labels agree with the consensus on 86\% ($\kappa = 0.62$), and the LLM is right for 99\% of the claims it keeps but only 55\% of those it removes, so the drops under the relevance filter are an upper bound.\looseness=-1

The limitations discussed above underscore that both long-form and short-form scores serve as mere proxies for factuality rather than direct measures. Long-form evaluation assesses the consistency of extracted claims with retrieved evidence, whereas short-form evaluation relies on a judge's assessment of agreement with a reference answer. Consequently, the lack of improvements on these metrics does not rule out gains in factuality in other ways. Future comparisons should use a common evaluation, report relevance-aware scores with claim counts and abstention rates, and release implementations matching the paper.\looseness=-1

\section{Conclusions}
\label{sec:conclusion}

In this paper, we examined the reproducibility and comparability of five factuality alignment methods, evaluating each under its original experimental setup and under a shared setup with a common base model, judge, and verifier.
None fully reproduced its published results. Under the shared setup, no method significantly outperformed the untrained base model, except for an artifact associated with one-word answers and a small gain after relevance filtering. Our analysis further suggests that the rewards favor length, format, or brevity rather than the factuality measured by the benchmarks, and that on-policy and offline training perform alike.
We also found that evaluation choices, including relevance filtering, evidence sources, recall normalization, and the judge, affected scores more than the methods themselves.
These findings highlight the need for shared evaluation settings, relevance-aware metrics accompanied by claim counts and abstention rates, and released implementations consistent with the reported experiments.
Future work could consider additional methods and larger models, expand the human evaluation to more annotators and claims, and investigate rewards that credit only relevant, supported claims.\looseness=-1

\section*{Limitations}
Our common evaluation uses Llama-3.1-8B-Instruct with LoRA adapters, four English datasets, and smaller training budgets than the original papers. Results may differ with larger models, longer training, or full fine-tuning. Each method was trained once, leaving variance across seeds unmeasured.

We document substitutions for unavailable artifacts, but cannot fully separate their effects from those of the methods. We also cannot inspect the training conditions of released checkpoints. Only one of the five author groups contacted had replied at the time of writing, leaving some discrepancies unresolved.

All scores rely on automatic claim extraction, web-based verification, commercial judges whose snapshots may change, and relevance labels from one LLM. Our human audit covers 300 claims checked by two annotators. Reference verification covers biographies only, as other datasets lack reference text. The fixed recall normalizer is one of several reasonable choices.

\section*{Ethical Considerations}
All data come from public benchmarks. The released biography generations contain false claims about real people and are intended for factuality research, not as sources of biographical information.

We follow the models' licenses and release code, configurations, and adapters rather than full weights. The appendix reports GPU usage and the code repository estimated API costs; billed amounts were not recorded. The human audit was carried out by two of the authors.

Our reward-hacking results show that higher automatic scores need not imply more truthful answers. These scores alone should not certify a system's factual reliability.

\section*{Use of AI Assistants}
We used Claude Code to help run experiments, to generate scripts for analyzing results and producing LaTeX plots and diagrams. We also used Opus 5.5 to rephrase parts of the paper to improve readability in addition to fixing typos and grammatical issues. All AI-assisted work was carried out under the authors' supervision. The authors take full responsibility for the resulting code, experimental procedures, analyses, and claims made in the paper, including any errors that remain.

\bibliography{references}

\appendix
\renewcommand{\topfraction}{0.9}
\renewcommand{\bottomfraction}{0.8}
\renewcommand{\textfraction}{0.07}
\renewcommand{\floatpagefraction}{0.75}
\renewcommand{\dbltopfraction}{0.9}
\renewcommand{\dblfloatpagefraction}{0.75}
\setcounter{topnumber}{4}
\setcounter{bottomnumber}{3}
\setcounter{totalnumber}{6}
\setcounter{dbltopnumber}{3}
\label{sec:appendix}

\section{Reproductions in the Original Setups}
\label{app:reproductions}

This appendix gives, for each method, the setup we reproduced, the deviations from the paper, and the results under the paper's own metric. The summary is in Section~\ref{sec:origsetup} and the published-versus-ours comparison in Table~\ref{tab:papervsours}.

\subsection{\flame{}}
\label{app:flame}

The paper fine-tunes Llama-2 70B in four variants, SFT and SFT followed by DPO as the standard pipeline, and their factuality-aware counterparts \textsc{Sft}$^{\text{f}}$ and \flame{}, which is \textsc{Sft}$^{\text{f}}$ followed by DPO$^{\text{f}}$. We retrain all four with data rebuilt from the published sources, 3,200 Open Assistant pairs for instruction following and 1,630 judge examples for evaluation fine-tuning, 20K self-generated instructions, and preference data ranked by an instruction-following and a factuality reward model. Full fine-tuning of the 70B model runs with DeepSpeed ZeRO-3 and optimizer offload on two H200 GPUs, at about 260 seconds per SFT step and 500 per DPO step. The paper mixes its instruction and evaluation fine-tuning data in the SFT stage, whereas our \textsc{Sft}$^{\text{f}}$ set is built from the 3,200 instruction-following prompts alone, with the fact-seeking ones answered by Llama-2-70B samples. Three components are substituted. The paper's reward models are not released, so the instruction-following reward is the 70B SFT model judging itself and the factuality reward is the recovered Inst-LLaMA checker. The factuality of open-ended instructions is scored with the same pipeline over an open retriever rather than with article-based FActScore. The AlpacaEval judge is \texttt{gpt-4-turbo}, since the paper's \texttt{gpt-4-1106-preview} is retired.

Table~\ref{tab:main} shows that standard DPO lowers biography FActScore from 53.9 to 51.6, while factuality-aware DPO lowers it from 54.6 to 54.2. \flame{} exceeds SFT+DPO by 2.6 points on biographies and 2.5 on Alpaca-Fact, against a published biography gain of 5.6. \textsc{Sft}$^{\text{f}}$ also improves TruthfulQA BLEU and ROUGE-L over SFT, but \flame{} shows no FAVA advantage with our open retriever.

The helpfulness claim does not reproduce: our AlpacaEval win rate is 37.7\%, or 38.6\% length-controlled, against the reported 51.2\%. Substituted reward models may contribute to this gap, but one training run and a newer judge do not allow us to isolate the cause.

\begin{table}[H]

\centering
\small
\begin{tabular}{lcccc}
\toprule
Llama-2 70B & Bio & A-Fact & FAVA & Win\% \\
\midrule
SFT & 53.9 & 44.3 & 54.7 & 26.1 \\
SFT + DPO & 51.6 & 43.7 & 51.5 & 50.0 \\
\textsc{Sft}$^{\text{f}}$ & 54.6 & 45.5 & 54.2 & 19.3 \\
\flame{} & 54.2 & 46.2 & 50.9 & 37.7 \\
\bottomrule
\end{tabular}
\caption{\flame{} at 70B: FActScore on 183 biographies, 253 Alpaca-Fact prompts, and FAVA; AlpacaEval win rate against SFT+DPO using \texttt{gpt-4-turbo}. Published biography scores are 44.7, 42.3, 49.5, and 47.9 for the four rows.}
\label{tab:main}
\end{table}

\begin{table*}[t]
  \centering

  \footnotesize
  \setlength{\tabcolsep}{4pt}
  \begin{tabular}{@{}l c >{\raggedright\arraybackslash}p{0.20\textwidth} c >{\raggedright\arraybackslash}p{0.16\textwidth} >{\raggedright\arraybackslash}p{0.29\textwidth}@{}}
  \toprule
  \textbf{Method} & \textbf{Code} & \textbf{Training data} & \textbf{Checkpoints} & \textbf{Reward or judge} & \textbf{Released code vs.\ paper} \\
  \midrule
  \flame & \xmark & prompts only; SFT and DPO data are self-generated & \xmark & reward models \xmark & no code to compare \\
  \addlinespace
  \factalign & \cmark & self-generated; general-domain mixture? & \cmark & SAFE pipeline \cmark & evaluates a LongFact variant regenerated in 2026, which offsets every absolute number by about 19 points \\
  \addlinespace
  \knowrl & \cmark & \cmark & \cmark & API judge, snapshot drifts & cold-start stage via LLaMA-Factory with the released data; the released RL configuration sets learning rate $10^{-4}$ and rank 128 where the paper gives $10^{-5}$ and 256, with the base model left as a placeholder; the fact reward receives the response without its opening reasoning tag \\
  \addlinespace
  \factgrpo & \cmark & SFT data \cmark; RL prompts recovered from the DPO pairs (3{,}706 of 4k) & ? & programmatic (\veriscore) \cmark & $\lambda_{\text{len}} = 0$ in the released configuration \\
  \addlinespace
  \faith & \cmark & \cmark, with the estimator's predicted state in every prompt & ? & reward model trained from released code & one epoch for the reward model instead of two and 0.45 PPO epochs instead of about two in the Llama-3 configurations; the rectifier ships a Mistral configuration but no training data (provided by the authors on request) and no Llama-3 configuration \\
  \bottomrule
  \end{tabular}
    \caption{What each method releases and where the released code departs from its paper. \cmark\ released, \xmark\ not released, \emph{partial} when only a part is usable, and ? when it is unclear.}
\label{tab:availability}
  \end{table*}

\subsection{\factalign{}}
\label{app:factalign}

The paper runs three iterations of fine-grained KTO, mixing newly generated responses with general-domain preference data, and evaluates SAFE $\text{F1}@100$ on LongFact. We test both the released checkpoint and retraining.

We score the checkpoint and Llama-3-8B-Instruct together on 190 prompts using one judge session and retriever state. Repeated evaluation of one model on 64 prompts gives a 1.51-point difference (CI: $-2.05$ to $+5.04$), encompassing our initial unpaired gain of $+2.72$. The paired gain is $+2.68$ (CI: $+1.28$ to $+4.15$; $p=.0003$). Claim counts remain near 46, while unsupported claims fall by 19\% and supported claims rise by 6\% (Table~\ref{tab:factalign}). Replacing web search with the rebuilt December 2021 Wikipedia index of 33 million passages gives a $+3.13$-point gain.

Our three-iteration gemma-2b run uses the paper's effective batch size of 16 but omits the general-domain mixture. $\text{F1}@100$ falls from 0.428 to 0.401, a 2.64-point loss (CI: $-4.49$ to $-0.90$); the first iteration scores 0.366. This agrees in direction with the paper's no-mixture ablation, which falls from 61.74 to 61.33.

Across 570 generations per model, mean length falls from 705 to 593 words, and responses repeating a sentence at least three times fall from 33\% to 19\%. Claim density nevertheless rises from 7.7 to 10.0 per 100 words, with 1--253 claims per response. Unsupported claims grow faster than supported ones. Beyond 100 supported claims, recall cannot increase, while unsupported claims can still lower precision.

Our LongFact version was regenerated in 2026 rather than 2024, shifting absolute scores by about 19 points on identical weights. We therefore compare models within the same evaluation.

\begin{table}[H]
\centering

\small
\begin{tabular}{@{}lcc@{}}
\toprule
& Llama-3-8B-Inst. & \factalign{} \\
\midrule
claims          & 46.00 & 45.22 \\
supported       & 30.54 & \textbf{32.39} \\
not supported   & 14.71 & \textbf{11.98} \\
precision       & 0.675 & \textbf{0.730} \\
$\text{F1}@100$ & 0.4145 & \textbf{0.4413} \\
\bottomrule
\end{tabular}
\caption{\factalign{}'s released checkpoint against its base,
both rated in one paired SAFE pass over 190 prompts. Counts are per
response.}
\label{tab:factalign}
\end{table}

\subsection{\knowrl{}}
\label{app:knowrl}

The paper applies cold-start SFT followed by GRPO. Its reward combines format ($\pm1$), correctness ($+2$ correct, $-1$ incorrect, $+1$ refusal), and FActScore over reasoning for incorrect, non-refused answers. The repository provides cold-start data and refers to LLaMA-Factory, but leaves the RL model path unspecified. Its learning rate and adapter rank are $10^{-4}$ and 128, against $10^{-5}$ and 256 in the paper.

Our first four runs reach 97\%--100\% refusals in the final fifth of training, with completion length falling from 344 to 89 tokens (Table~\ref{tab:knowrl_runs}). Two use released hyperparameters, with and without cold start; two use the paper's hyperparameters without cold start. Cold-start SFT followed by 150 GRPO steps at $10^{-5}$ and rank 256 instead keeps refusals near 0.2\% and reward near $-0.64$. Cold start alone does not prevent collapse at the released learning rate.

The correctness term gives an attempt expected reward $3p-1$, against $+1$ for refusal, favoring refusal when $p<2/3$. A separate formatting error omits the opening reasoning tag from the factuality reward's input, reducing extracted facts from 11.8 to 5.3 per response and leaving two fifths without facts. Repairing the tag restores the signal but does not prevent collapse. Our first cold-start model reduces refusals from 22.7\% to 3.7\% on 300 questions.

The checkpoint's SimpleQA gain reproduces: 1.92 to 4.56 in the paper, and 2.00 to 5.71 on our 500 items (Table~\ref{tab:knowrl_sqa}). However, accuracy excludes refusals. Rejection rises from 10.2\% to 93.0\%, and correct answers fall from nine to two. The higher score therefore concerns a much smaller set of attempted questions.

\begin{table}[H]
\centering

\small
\setlength{\tabcolsep}{4pt}
\begin{tabular}{p{3cm}rrrrr}
\toprule
Run & 1 & 2 & 3 & 4 & 5 \\
\midrule
\knowrl{}, end to end & \textbf{0.2} & \textbf{0.2} &
  \textbf{0.1} & \textbf{0.2} & \textbf{0.2} \\
\textsc{KnowRL-Zero}, paper hyperparameters & 7.2 & 14.2 & 33.7 & 93.1 & 96.8 \\
\quad + repaired fact input & 7.4 & 15.2 & 82.2 & 98.3 & 98.3 \\
released hyperparameters, raw model & 65.2 & 99.6 & 99.9 & 100.0 & 100.0 \\
released hyperparameters, after cold start & 0.9 & 15.7 & 85.2 & 99.2 & 98.4 \\
\bottomrule
\end{tabular}
\caption{\knowrl{} refusal rate in percent by fifth of training. The method as specified does not abstain, whereas every run at the released learning rate or without the paper's cold start collapses, and repairing the input of the fact reward does not prevent the collapse, which locates the cause in the cold-start stage and the learning rate rather than in the fact signal.}
\label{tab:knowrl_runs}
\end{table}

\begin{table}[H]
\centering

\small
\begin{tabular}{lrr}
\toprule
 & Base distill & \knowrl{} \\
\midrule
Reported accuracy & 1.92 & 4.56 \\
\hspace{1em}our measurement & 2.00 & \textbf{5.71} \\
\midrule
Rejected & 10.2\% & \textbf{93.0\%} \\
Correct answers, of 500 & 9 & \textbf{2} \\
\bottomrule
\end{tabular}
\caption{The released \knowrl{} checkpoint on 500 SimpleQA items under the authors' protocol. The reported gain reproduces, but rejection rises from 10\% to 93\% and the number of correctly answered questions falls.}
\label{tab:knowrl_sqa}
\end{table}

\subsection{\factgrpo{}}
\label{app:factgrpo}

Mean rewards across 80-step blocks are 0.704, 0.704, 0.686, 0.741, and 0.718: a net rise of 0.014 against batch fluctuations up to 0.9. No malformed response occurs in 400 steps. Before evaluation, we predicted a gain far below the published result. Precision rises by 1.1 points on biographies and 2.8 on Alpaca-Fact, roughly a tenth of the published gains, while detail level falls (Table~\ref{tab:chenresults}).

Although $S/(1+T)$ rewards more claims at fixed positive precision, our reduced-budget run does not recover the reported joint improvement. The required format appears in 83\% of biography and 60\% of Alpaca-Fact responses, but only 6\% of SimpleQA and under 1\% of TruthfulQA responses. SimpleQA answers are almost five times the base model's length. Most short-form outputs would therefore incur the model's own format penalty.

\paragraph{Reward ablation.}
The first run omitted the relevance term. We repeated training with the full reward and unchanged data, budget, and adapter; all common benchmark tables use this rerun. The runs receive identical grades on 80\% of TruthfulQA and 81\% of SimpleQA items, with no significant difference under McNemar's exact test (Tables~\ref{tab:factgrpo_ablation_short} and~\ref{tab:factgrpo_ablation_long}). Long-form scores differ by less than two points. The judge prefers the rerun's responses to the base model's in about one third of cases, contributing .03 to a mean reward near .3. We find no measurable benefit from this term under our budget.

\begin{table}[H]
\centering
\small
\begin{tabular}{@{}l cccc@{}}
\toprule
Method & Acc. & Abst. & Acc$\mid$att. & F \\
\midrule
\multicolumn{5}{@{}l}{SimpleQA ($n = 1{,}000$)} \\
without the relevance term & .053 & .401 & .088 & .066 \\
full reward                & .052 & .414 & .089 & .066 \\
\addlinespace
\multicolumn{5}{@{}l}{TruthfulQA ($n = 817$)} \\
without the relevance term & .237 & .355 & .368 & .289 \\
full reward                & .231 & .368 & .366 & .284 \\
\bottomrule
\end{tabular}

\caption{Short-form results of the two \factgrpo{} runs on the common benchmark, graded by \texttt{gpt-4o}. The first run was trained without the relevance term of the reward, and the rerun with the full reward is the variant used in the main tables.}
\label{tab:factgrpo_ablation_short}
\end{table}

\begin{table}[H]
\centering
\caption{Our \factgrpo{} run against the published configuration of \citet{Chen:2026:ICML}. The last three rows are the deviations that matter.}
\label{tab:chenconfig}
\small
\setlength{\tabcolsep}{4pt}
\begin{tabularx}{\columnwidth}{@{}ll>{\raggedright\arraybackslash}X@{}}
\toprule
& \textbf{Paper} & \textbf{Ours} \\
\midrule
Base model     & Llama-3.1-8B-Inst. & same \\
SFT data       & 3k synthetic  & their released 3k \\
RL prompts     & 4k synthetic  & 3{,}706 recovered \\
Rollouts       & 4             & 4 \\
Learning rate  & 1e-6          & 1e-6 \\
Objective      & Dr.\ GRPO     & Dr.\ GRPO \\
\midrule
Training       & ${\sim}3{,}706$ steps & \textbf{400 steps (11\%)} \\
Fine-tuning    & full          & \textbf{LoRA $r{=}32$} \\
Hardware       & 40 H100       & \textbf{1 H100 ${+}$ 1 A100,}\newline\textbf{3 A100 (rerun)} \\
\bottomrule
\end{tabularx}
\end{table}

\begin{table}[H]
\centering
\caption{\veriscore{} precision ($P$) and detail level ($S$) on the long-form benchmarks, base model against our 400-step run, with the published effect for reference.}
\label{tab:chenresults}
\small
\setlength{\tabcolsep}{4pt}
\begin{tabular}{@{}l cc cc@{}}
\toprule
& \multicolumn{2}{c}{Biography} & \multicolumn{2}{c}{Alpaca-Fact} \\
\cmidrule(lr){2-3}\cmidrule(lr){4-5}
& $P$ & $S$ & $P$ & $S$ \\
\midrule
Base            & 35.8 & 12.1 & 61.9 & 20.0 \\
${+}$ our GRPO  & \textbf{36.9} & 10.9 & \textbf{64.7} & 18.1 \\
\addlinespace
Paper, base     & 25.9 & 10.3 & 60.2 & 30.7 \\
Paper, ${+}$GRPO & \textbf{47.1} & \textbf{14.8} & \textbf{79.7} & \textbf{36.9} \\
\bottomrule
\end{tabular}
\end{table}

\begin{table*}[t]
\centering
\caption{\factgrpo{} with and without the relevance reward, evaluated on shared prompts before and after relevance filtering. $K$ matches Table~\ref{tab:factuality}.}
\label{tab:factgrpo_ablation_long}
\small
\begin{tabular}{lrrrrrrrrr}
\toprule
 & & \multicolumn{4}{c}{VS} & \multicolumn{4}{c}{VS+Rel} \\
\cmidrule(lr){3-6} \cmidrule(lr){7-10}
Method & \#Claims & \#Sup. & P & R & F1 & \#Sup. & P & R & F1 \\
\midrule
\multicolumn{10}{l}{\textit{Biography ($n = 488$, $K = 20$)}} \\
without the relevance term & 25.6 & 10.5 & 0.360 & 0.480 & 0.399 & 9.4 & 0.316 & 0.438 & 0.357 \\
\midrule
full reward                & 25.4 & 10.1 & 0.342 & 0.466 & 0.384 & 9.1 & 0.301 & 0.424 & 0.343 \\
\midrule
\multicolumn{10}{l}{\textit{Alpaca-Fact ($n = 246$, $K = 19$)}} \\
without the relevance term & 26.5 & 18.2 & 0.631 & 0.699 & 0.638 & 13.4 & 0.471 & 0.573 & 0.491 \\
\midrule
full reward                & 25.6 & 17.7 & 0.625 & 0.681 & 0.629 & 13.3 & 0.472 & 0.570 & 0.492 \\
\bottomrule
\end{tabular}

\end{table*}

\subsection{\faith{}}
\label{app:faith}

\faith{} trains a knowledge-state estimator, SFT model, reward model, and state-conditioned PPO policy, followed by a retrieval-based rectifier. The artifact includes four datasets, a trainer, and stage-specific configurations.

We train the first four stages on \texttt{Meta-Llama-3-8B}: 3 epochs each for the estimator and SFT, 2 for the reward model, and 2 PPO epochs over 10,000 of 93,904 prompts, about 310 steps or one fifth of a full-data pass. We follow the paper's epoch counts rather than the released settings of 1 for the reward model and 0.45 for PPO.

We omit the rectifier because its data and Llama-3 configuration were unavailable during our runs. The authors later supplied the data and confirmed that the configuration was lost. We therefore compare against the no-rectifier ablation. Our estimator replaces the authors' stored evaluation states; using those states changes accuracy by at most 0.7 points, with 87.5\% state agreement (Table~\ref{tab:faithstate}).

PPO adds 0.33 points in mean precision over questions classified as known, against 1.20 for the published ablation and about ten for the full system (Table~\ref{tab:faith}). Our SFT scores match the reported levels on TriviaQA and NQ-Open but fall short on SciQ and WebQuestions. Reward changes little, from $-0.37$ at step 9 to $-0.34$ at step 309.

Questions classified as known are answered correctly more than twice as often as unknown questions on every dataset. Yet refusal rates are 0.000 on three datasets and 0.0003 on the fourth, despite 17\%--56\% of questions being classified as unknown. Truthfulness, which credits correct refusals, is therefore effectively accuracy.

\begin{table}[H]
\centering
\caption{\faith{} with predicted (Est.) or released, precomputed (Read) knowledge states. Agree measures state agreement. Both models use \texttt{Meta-Llama-3-8B} (Section~\ref{sec:original}); Est. follows Section~3.4 of the original paper.}
\label{tab:faithstate}
\small
\setlength{\tabcolsep}{5pt}
\begin{tabular}{@{}l cc cc c@{}}
\toprule
& \multicolumn{2}{c}{SFT} & \multicolumn{2}{c}{SFT $+$ PPO} & \\
\cmidrule(lr){2-3}\cmidrule(lr){4-5}
Dataset & Est. & Read & Est. & Read & Agree \\
\midrule
NQ-Open        & .297 & .303 & .295 & .298 & 83.2\% \\
SciQ           & .585 & .584 & .598 & .591 & 84.1\% \\
TriviaQA       & .662 & .662 & .659 & .658 & 90.1\% \\
WebQuestions   & .342 & .341 & .347 & .346 & 82.6\% \\
\midrule
overall        & \multicolumn{4}{c}{} & 87.5\% \\
\bottomrule
\end{tabular}
\end{table}

\begin{table}[H]
\centering
\caption{\faith{} on its own four validation sets, scored in one
harness. Precision is over the questions the estimator predicts as
known. The last column is the gain of the paper's full system on Llama-3-8B, whereas our delta compares our own SFT stage with our own PPO model, which corresponds to the paper's ablation without the retrieval stage, reported in Table~\ref{tab:papervsours}.}
\label{tab:faith}
\small
\setlength{\tabcolsep}{4pt}
\begin{tabular}{@{}l cc c c@{}}
\toprule
& \multicolumn{2}{c}{Ours, precision} & & Full system \\
\cmidrule(lr){2-3}
Dataset & \textsc{Sft} & \faith{} & $\Delta$ & $\Delta$ \\
\midrule
TriviaQA      & 74.15 & 73.56 & $-0.60$ & $+13.39$ \\
SciQ          & 65.74 & 67.13 & $+1.39$ & $+8.43$ \\
NQ-Open       & 42.51 & 42.57 & $+0.06$ & $+16.72$ \\
WebQuestions  & 43.27 & 43.74 & $+0.47$ & $+1.53$ \\
\midrule
Mean          &       &       & $+0.33$ & $+10.02$ \\
\bottomrule
\end{tabular}
\end{table}

\paragraph{Why \faith{} produces short answers.}
The reward model's containment label accepts a prediction contained in the gold answer, or vice versa. This can credit incomplete answers, but does not by itself establish a preference for brevity. The short-answer SFT data offer a more direct explanation: mean biography length falls from 274 words in the base model to 3 after SFT and remains 3 after PPO. Adding the knowledge state does not restore long responses (Table~\ref{tab:faith_state_long}). Since \faith{} is presented as general factuality alignment, this behavior matters beyond its short-answer training tasks.

\section{Additional Common-Benchmark Results}
\label{app:common}

This section gives the common-benchmark results that Sections~\ref{sec:reward} and~\ref{sec:sftdecomp} refer to.

\subsection{Supervised Stages Alone}

Tables~\ref{tab:sftonly} and~\ref{tab:longdecomp} give the short-form and long-form results of each method's supervised stage alone, which Section~\ref{sec:sftdecomp} compares with the full pipelines in Tables~\ref{tab:shorttrack} and~\ref{tab:factuality}.

\begin{table}[H]
\centering
\caption{SFT stage results for short-form evaluation.}
\label{tab:sftonly}
\small
\setlength{\tabcolsep}{4pt}
\begin{tabular}{@{}l cccc@{}}
\toprule
Method & Acc. & Abst. & Acc$\mid$att. & F \\
\midrule
\multicolumn{5}{@{}l}{SimpleQA ($n = 1{,}000$)} \\
base (untrained) & .034 & .812 & .181 & .057 \\
\flame           & .029 & \textbf{.860} & \textbf{.207} & .051 \\
\knowrl          & .029 & .831 & .172 & .050 \\
\factgrpo        & \textbf{.054} & .416 & .092 & \textbf{.068} \\
\faith           & .046 & .001 & .046 & .046 \\
\addlinespace
\multicolumn{5}{@{}l}{TruthfulQA ($n = 817$)} \\
base (untrained) & .416 & .108 & .466 & .440 \\
\flame           & \textbf{.431} & .100 & \textbf{.479} & \textbf{.454} \\
\knowrl          & .401 & .131 & .462 & .430 \\
\factgrpo        & .236 & \textbf{.365} & .372 & .289 \\
\faith           & .165 & .029 & .170 & .168 \\
\bottomrule
\end{tabular}

\end{table}

\begin{table}[H]
\centering
\caption{Long-form results of the supervised stages alone, scored in the same \veriscore{} configuration and with the same fixed $K$ as Table~\ref{tab:factuality}.}
\label{tab:longdecomp}
\small
\setlength{\tabcolsep}{5pt}
\begin{tabular}{@{}l ccc cc@{}}
\toprule
Method & Sent. & $S$ & $U$ & Prec. & $\text{F1}@K$ \\
\midrule
\multicolumn{6}{@{}l}{Biography ($n = 491$, $K = 20$)} \\
base (untrained) & 13.6 & 12.02 & 15.60 & .351 & .370 \\
\flame           & 11.6 & 10.30 & 12.35 & .334 & .349 \\
\knowrl          & 16.0 & 11.33 & 19.31 & .270 & .316 \\
\factgrpo        & 12.7 & 10.33 & 15.15 & .353 & .394 \\
\faith           & 1.0 & 0.39 & 0.21 & .373 & .037 \\
\addlinespace
\multicolumn{6}{@{}l}{Alpaca-Fact ($n = 247$, $K = 19$)} \\
base (untrained) & 21.9 & 19.79 & 9.38 & .618 & .613 \\
\flame           & 19.6 & 17.32 & 8.36 & .594 & .573 \\
\knowrl          & 23.3 & 19.59 & 10.79 & .584 & .594 \\
\factgrpo        & 14.8 & 17.10 & 8.37 & .616 & .616 \\
\faith           & 1.0 & 0.13 & 0.03 & .103 & .012 \\
\bottomrule
\end{tabular}

\end{table}

\subsection{\faith{} With and Without the Knowledge State}

Tables~\ref{tab:faith_state_short} and~\ref{tab:faith_state_long} compare \faith{} with and without the estimator-predicted knowledge state in the prompt, for the supervised stage and the full pipeline. The state lowers accuracy, precision, and $\text{F1}@K$ slightly, and the full pipeline barely differs from its supervised stage. The main results use the variant with the state, as the paper prescribes.

\begin{table}[H]
\centering
\caption{Short-form results of \faith{} with and without the estimator-predicted knowledge state in the prompt, for the supervised stage and the full pipeline.}
\label{tab:faith_state_short}
\small
\setlength{\tabcolsep}{4pt}
\begin{tabular}{@{}l cccc@{}}
\toprule
Method & Acc. & Abst. & Acc$\mid$att. & F \\
\midrule
\multicolumn{5}{@{}l}{SimpleQA ($n = 1{,}000$)} \\
\faith{} SFT          & .055 & \textbf{.001} & .055 & .055 \\
\faith{} SFT + state  & .046 & \textbf{.001} & .046 & .046 \\
\faith{} full         & \textbf{.058} & \textbf{.001} & \textbf{.058} & \textbf{.058} \\
\faith{} full + state & .049 & .000 & .049 & .049 \\
\addlinespace
\multicolumn{5}{@{}l}{TruthfulQA ($n = 817$)} \\
\faith{} SFT          & .192 & .020 & .196 & .194 \\
\faith{} SFT + state  & .165 & \textbf{.029} & .170 & .168 \\
\faith{} full         & \textbf{.193} & .021 & \textbf{.198} & \textbf{.195} \\
\faith{} full + state & .170 & \textbf{.029} & .175 & .173 \\
\bottomrule
\end{tabular}

\end{table}

\begin{table}[H]
\centering
\caption{Long-form results of the same four \faith{} variants, with the fixed $K$ of Table~\ref{tab:factuality}.}
\label{tab:faith_state_long}
\small
\setlength{\tabcolsep}{4pt}
\begin{tabular}{@{}l ccc cc@{}}
\toprule
Method & Sent. & $S$ & $U$ & Prec. & $\text{F1}@K$ \\
\midrule
\multicolumn{6}{@{}l}{Biography ($n = 495$, $K = 20$)} \\
\faith{} SFT          & 1.0 & 0.54 & 0.24 & .446 & .050 \\
\faith{} SFT + state  & 1.0 & 0.39 & 0.21 & .374 & .037 \\
\faith{} full         & 1.0 & 0.52 & 0.22 & .444 & .048 \\
\faith{} full + state & 1.0 & 0.39 & 0.21 & .377 & .037 \\
\addlinespace
\multicolumn{6}{@{}l}{Alpaca-Fact ($n = 253$, $K = 19$)} \\
\faith{} SFT          & 1.2 & 0.25 & 0.06 & .132 & .023 \\
\faith{} SFT + state  & 1.0 & 0.12 & 0.04 & .101 & .012 \\
\faith{} full         & 1.2 & 0.24 & 0.05 & .134 & .022 \\
\faith{} full + state & 1.0 & 0.12 & 0.03 & .099 & .012 \\
\bottomrule
\end{tabular}

\end{table}

\subsection{Cross-Reward Scores}
\label{app:crossreward}

Table~\ref{tab:crossreward} gives the mean reward of each method's generations under each method's reward, which Section~\ref{sec:reward} discusses.

\begin{table}[H]
\centering
\caption{Mean reward of each method's generations under each method's reward, as the difference from the base model's generations, whose absolute value is in the first row. Rewards are per thousand words (\flame{}, \factalign{}, \faith{}) or per response (\knowrl{} on short-form and \factgrpo{} on long-form prompts). Bold marks a method under its own reward, and † a significant difference from the base row (paired bootstrap, Holm-corrected $p < 0.05$). \faith{} generations are left out, since one-word answers make per-word values not comparable.}
\label{tab:crossreward}
\footnotesize
\setlength{\tabcolsep}{3pt}
\resizebox{\columnwidth}{!}{\begin{tabular}{@{}lrrrrr@{}}
\toprule
 & \multicolumn{5}{c}{Reward} \\
\cmidrule(l){2-6}
Generations      & \flame & \factalign & \knowrl & \factgrpo & \faith \\
\midrule
base (untrained) & $1.13$ & $-1.15$ & $-0.392$ & $-1.000$ & $3.85$ \\
\midrule
\flame           & {\boldmath$-0.09$}\rlap{$^{\dagger}$} & $-0.25$\rlap{$^{\dagger}$} & $+0.050$\rlap{$^{\dagger}$} & $0.000$ & $+2.16$\rlap{$^{\dagger}$} \\
\factalign       & $-0.01$ & {\boldmath$+0.05$}\rlap{$^{\dagger}$} & $+0.026$ & $0.000$ & $+0.03$ \\
\knowrl          & $-0.12$\rlap{$^{\dagger}$} & $+0.03$ & {\boldmath$+0.004$} & $0.000$ & $-0.32$ \\
\factgrpo        & $-0.40$\rlap{$^{\dagger}$} & $-2.24$\rlap{$^{\dagger}$} & $-0.349$\rlap{$^{\dagger}$} & {\boldmath$+1.028$}\rlap{$^{\dagger}$} & $-3.97$\rlap{$^{\dagger}$} \\
\bottomrule
\end{tabular}
}
\end{table}

\subsection{\factgrpo{} Reward Hacking}

The reward rises with the number of supported but irrelevant claims because its main term, $S/(1+T)$, counts every supported claim and the relevance term contributes little, so filler claims raise the reward of the rerun as they did in the first run (Figure~\ref{fig:factgrpo_reward_hacking} in Section~\ref{sec:reward}).

\section{Evaluation Analysis}
\label{app:evaluation}

This section gives the analyses of the evaluation that Section~\ref{sec:rq4} refers to.

\subsection{Open LLM Judge}

Table~\ref{tab:openjudge} repeats the short-form evaluation with the open \texttt{gpt-oss-120b} judge in place of \texttt{gpt-4o}. The open judge grades TruthfulQA more leniently, but the large drops of \factgrpo{} and \faith{} remain.

\begin{table}[H]
\centering
\caption{The short-form evaluation under both judges. \emph{Acc.} is accuracy over
all questions, \emph{F} the harmonic mean of accuracy and
accuracy-given-attempted. Absolute values shift systematically between
judges; only differences within a column are meaningful. Items that the open judge left ungraded, up to 58 of 817 on TruthfulQA, are excluded from its rates.}
\label{tab:openjudge}
\small
\setlength{\tabcolsep}{4pt}
\begin{tabular}{@{}l cc cc@{}}
\toprule
& \multicolumn{2}{c}{\texttt{gpt-4o}} & \multicolumn{2}{c}{\texttt{gpt-oss-120b}} \\
\cmidrule(lr){2-3}\cmidrule(lr){4-5}
Method & Acc. & F & Acc. & F \\
\midrule
\multicolumn{5}{@{}l}{SimpleQA ($n = 1{,}000$)} \\
base (untrained) & .034 & .057 & .030 & .051 \\
\flame           & .032 & .056 & .030 & .052 \\
\factalign       & .032 & .054 & .029 & .049 \\
\knowrl          & .032 & .054 & .031 & .052 \\
\factgrpo        & \textbf{.052} & \textbf{.066} & \textbf{.053} & \textbf{.067} \\
\faith           & .049 & .049 & .048 & .049 \\
\addlinespace
\multicolumn{5}{@{}l}{TruthfulQA ($n = 817$)} \\
base (untrained) & .416 & .440 & .466 & .501 \\
\flame           & .424 & .448 & \textbf{.471} & .505 \\
\factalign       & \textbf{.435} & \textbf{.460} & .469 & \textbf{.506} \\
\knowrl          & .403 & .429 & .459 & .502 \\
\factgrpo        & .231 & .284 & .294 & .361 \\
\faith           & .170 & .173 & .186 & .202 \\
\bottomrule
\end{tabular}

\end{table}

\begin{table*}[t]
\centering
\small
\setlength{\tabcolsep}{4pt}
\renewcommand{\arraystretch}{1.15}
\begin{tabularx}{\textwidth}{@{}>{\raggedright\arraybackslash}p{0.155\textwidth}Xcc@{}}
\toprule
\textbf{Question} & \textbf{Response} (\texttt{<answer>} span) & $R_{\mathrm{rel}}$ & $R_{\mathrm{fact}}$ \\
\midrule
Tell me a bio of Diego Suarez (garden designer).
& I'm not sure if I have any information on a garden designer by the name of Diego Suarez. However, I can provide some general information about \emph{garden designers} \ldots{} Garden designers are professionals who design and plan gardens, taking into account factors such as the climate, soil type, and desired aesthetic. \ldots{} Some well-known garden designers include Gertrude Jekyll, Russell Page, and Piet Oudolf. \ldots{} I apologize, but I don't have any information on a garden designer by the name of Diego Suarez.
& 0.00 & 0.93 \\
\addlinespace
Tell me a bio of Shaolin (humorist).
& I'm not sure if I have any information about a well-known humorist by the name of Shaolin. However, I can provide some general information about \emph{the Shaolin Temple in China} \ldots{} The Shaolin Temple is located in Dengfeng, Henan Province, China \ldots{} has a rich history that dates back to the 5th century, and it's been a major center of Buddhist learning \ldots{} particularly Shaolin Kung Fu \ldots
& 0.00 & 0.84 \\
\addlinespace
Tell me a bio of Alexandre Hohagen.
& Alexandre Hohagen is a name that combines a French first name with a German surname. \emph{The name Alexandre is derived from the Greek name Alexander}, which means ``defender of the people.'' The surname Hohagen is of German origin \ldots{} Unfortunately, I don't have any specific information about a person named Alexandre Hohagen. \ldots
& 0.00 & 0.75 \\
\addlinespace
Tell me a bio of Kubota Beisen.
& I'm not familiar with a well-known artist by the name of Kubota Beisen. \ldots{} However, I can provide some general information about \emph{Japanese art and artists from the same time period}. The Edo period (1603--1868) and the Meiji period (1868--1912) were significant periods in Japanese art history \ldots{} Some notable Japanese artists from these periods include Hokusai \ldots{} Hiroshige \ldots{} and Yokoyama Taikan \ldots
& 0.00 & 0.74 \\
\addlinespace
Tell me a bio of Rangi Te Kanawa.
& Rangi Te Kanawa is a New Zealand singer, but I couldn't find much information about their background \ldots{} you might be thinking of \emph{Kiri Te Kanawa}. Kiri Te Kanawa is a prominent New Zealand soprano \ldots{} She was born in 1944 in Gisborne, New Zealand \ldots{} and has since performed with many major opera companies \ldots{} including the Metropolitan Opera in New York and the Royal Opera in London. \ldots
& 0.00 & 0.67 \\
\bottomrule
\end{tabularx}
\caption{\factgrpo{} responses from the first run that drift away from the requested subject yet earn a high factuality reward (\texttt{<answer>} span only; \ldots{} marks omitted text). Italics mark the substituted subject: a profession, a homonym, the name's etymology, an art-historical period, and a similarly named person. $R_{\mathrm{rel}}$ is 0.1 only when an LLM judge prefers the response to the reference response, and 0 otherwise; $R_{\mathrm{fact}} = S/(T{+}1)$ over $T$ extracted claims, $S$ of them supported. Because $R_{\mathrm{rel}} \le 0.1$, verifiable but off-topic content dominates the total reward.}
\label{tab:rfact-dominates-rrel}
\end{table*}

\subsection{Filler and Hallucinated Claims}

\begin{figure}[H]
\centering
\begin{tikzpicture}
\begin{axis}[
  width=\columnwidth, height=0.62\columnwidth,
  xlabel={\# Filler claims},
  ylabel={Count},
  xmin=0, xmax=15,
  ymode=log, log basis y=10, ymin=0.7, ymax=700, ytick={1,10,100},
  unbounded coords=jump,
  xtick={0,5,...,30},
  axis lines*=left,
  axis line style={draw=black!45},
  tick style={draw=black!45},
  tick label style={font=\footnotesize, text=black!70},
  label style={font=\footnotesize},
  ymajorgrids=true,
  grid style={line width=0.2pt, draw=black!12},
  enlarge x limits=0.03,
  every axis plot/.append style={line width=0.9pt, mark size=1.2pt},
  legend columns=3,
  legend cell align=left,
  legend style={
    font=\scriptsize, draw=none, fill=none,
    at={(0.5,1.03)}, anchor=south,
    /tikz/every even column/.append style={column sep=0.6em},
  },
]
  \addplot[color=cBase,      dashed, mark=none]                  table[x=n_filler, y=base]      {\fillerclaims};
  \addlegendentry{Base}
  \addplot[color=cFactAlign, mark=*]                          table[x=n_filler, y=factalign] {\fillerclaims};
  \addlegendentry{\factalign}
  \addplot[color=cFactGRPO,  mark=square*]                    table[x=n_filler, y=factgrpo]  {\fillerclaims};
  \addlegendentry{\factgrpo}
  \addplot[color=cFaith,     mark=triangle*, mark size=1.6pt] table[x=n_filler, y=faith]     {\fillerclaims};
  \addlegendentry{\faith}
  \addplot[color=cFlame,     mark=diamond*,  mark size=1.6pt] table[x=n_filler, y=flame]     {\fillerclaims};
  \addlegendentry{\flame}
  \addplot[color=cKnowRL,    mark=pentagon*]                  table[x=n_filler, y=knowrl]    {\fillerclaims};
  \addlegendentry{\knowrl}
\end{axis}
\end{tikzpicture}
\caption{Distribution of the number of filler (supported but irrelevant) claims within long-form responses for the Biography dataset. The y-axis is logarithmic, and a missing point means a count of zero.}
\label{fig:filler_claim_distribution}
\end{figure}
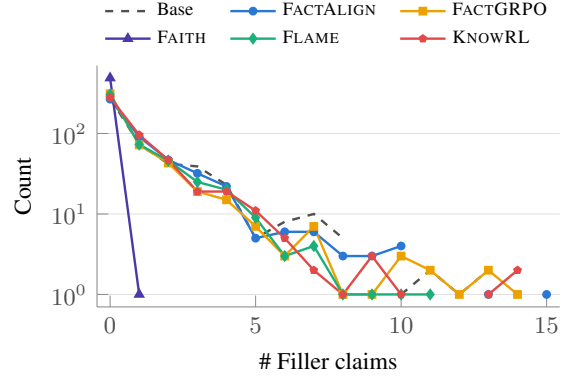

\begin{figure}[H]
\centering
\begin{tikzpicture}
\begin{axis}[
  width=\columnwidth, height=0.62\columnwidth,
  xlabel={\# Hallucinated claims},
  ylabel={Count},
  xmin=0, xmax=30,
  ymode=log, log basis y=10, ymin=0.7, ymax=700, ytick={1,10,100},
  unbounded coords=jump,
  xtick={0,5,...,25},
  axis lines*=left,
  axis line style={draw=black!45},
  tick style={draw=black!45},
  tick label style={font=\footnotesize, text=black!70},
  label style={font=\footnotesize},
  ymajorgrids=true,
  grid style={line width=0.2pt, draw=black!12},
  enlarge x limits=0.03,
  every axis plot/.append style={line width=0.9pt, mark size=1.2pt},
  legend columns=3,
  legend cell align=left,
  legend style={
    font=\scriptsize, draw=none, fill=none,
    at={(0.5,1.03)}, anchor=south,
    /tikz/every even column/.append style={column sep=0.6em},
  },
]
  \addplot[color=cBase,      dashed, mark=none]                  table[x=n_hallu, y=base]      {\hallucinatedclaims};
  \addlegendentry{Base}
  \addplot[color=cFactAlign, mark=*]                          table[x=n_hallu, y=factalign] {\hallucinatedclaims};
  \addlegendentry{\factalign}
  \addplot[color=cFactGRPO,  mark=square*]                    table[x=n_hallu, y=factgrpo]  {\hallucinatedclaims};
  \addlegendentry{\factgrpo}
  \addplot[color=cFaith,     mark=triangle*, mark size=1.6pt] table[x=n_hallu, y=faith]     {\hallucinatedclaims};
  \addlegendentry{\faith}
  \addplot[color=cFlame,     mark=diamond*,  mark size=1.6pt] table[x=n_hallu, y=flame]     {\hallucinatedclaims};
  \addlegendentry{\flame}
  \addplot[color=cKnowRL,    mark=pentagon*]                  table[x=n_hallu, y=knowrl]    {\hallucinatedclaims};
  \addlegendentry{\knowrl}
\end{axis}
\end{tikzpicture}
\caption{Distribution of the number of hallucinated (relevant but unsupported) claims within long-form responses for the Biography dataset. The y-axis is logarithmic, and a missing point means a count of zero.}
\label{fig:hallucinated_claim_distribution}
\end{figure}
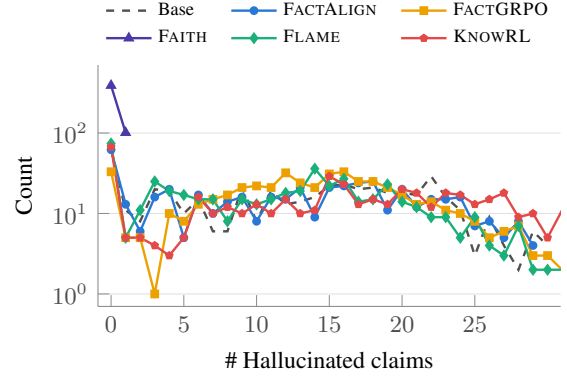

Figure~\ref{fig:filler_claim_distribution} shows the distribution of filler claims, supported but irrelevant, per biography response, and Figure~\ref{fig:hallucinated_claim_distribution} that of hallucinated claims, relevant but unsupported. Every method that answers at length has a long tail in both, whereas the one-word answers of \faith{} contain at most one claim of either kind.

\pgfplotstableread{
n_filler base factalign factgrpo faith flame knowrl
0         275       267      312   489   306    280
1          74        91       72     1    73     96
2          42        47       43     0    46     47
3          39        32       19     0    25     19
4          23        22       15     0    20     19
5           5         5        7     0     9     11
6           8         6        3     0     3      5
7          10         6        7     0     4      2
8           5         3        1     0     1      1
9           0         3        1     0     1      3
10          1         4        3     0     1      1
11          2         0        2     0     1      0
12          1         0        1     0     0      0
13          2         1        2     0     0      1
14          0         0        1     0     0      2
15          2         1        0     0     0      0
16          0         0        0     0     0      0
17          0         1        0     0     0      0
18          0         0        0     0     0      1
19          0         1        0     0     0      0
20          0         0        0     0     0      0
21          0         0        0     0     0      0
22          0         0        0     0     0      0
23          0         0        0     0     0      0
24          0         0        0     0     0      0
25          0         0        1     0     0      0
26          0         0        0     0     0      0
27          0         0        0     0     0      1
28          0         0        0     0     0      0
29          0         0        0     0     0      0
30          0         0        0     0     0      0
31          1         0        0     0     0      0
32          0         0        0     0     0      0
33          0         0        0     0     0      0
34          0         0        0     0     0      0
35          0         0        0     0     0      0
36          0         0        0     0     0      0
37          0         0        0     0     0      0
38          0         0        0     0     0      0
39          0         0        0     0     0      0
40          0         0        0     0     0      0
41          0         0        0     0     0      0
42          0         0        0     0     0      0
43          0         0        0     0     0      0
44          0         0        0     0     0      0
45          0         0        0     0     0      0
46          0         0        0     0     0      0
47          0         0        0     0     0      0
48          0         0        0     0     0      0
49          0         0        0     0     0      0
50          0         0        0     0     0      0
51          0         0        0     0     0      0
52          0         0        0     0     0      0
53          0         0        0     0     0      0
54          0         0        0     0     0      0
55          0         0        0     0     0      0
56          0         0        0     0     0      0
57          0         0        0     0     0      0
58          0         0        0     0     0      0
59          0         0        0     0     0      0
60          0         0        0     0     0      1
}\fillerclaims

\pgfplotstableread{
n_hallu base factalign factgrpo faith flame knowrl
0         63        62       33   389    74     67
1         11        13        5   101     5      5
2          8         6        5     0    11      5
3         20        16        1     0    25      4
4         19        20       10     0    19      3
5         10         5        8     0    17      5
6         15        17       13     0    15     16
7          6        10       15     0    15     10
8          6        14       17     0     8     12
9         21        16       21     0    15     10
10         8         8       22     0    13     13
11        19        16       21     0    15     10
12        13        15       32     0    18     15
13        14        21       24     0    19     10
14        16         9       21     0    36     11
15        22        21       31     0    22     29
16        23        22       33     0    27     23
17        20        24       25     0    14     13
18        21        25       25     0    15     15
19        18        11       21     0    23     13
20        18        20       17     0    14     20
21        17        12       13     0    12     18
22        29        15       14     0     9     12
23        16        15       11     0     9     18
24        11        16       10     0     5     17
25         3         7        8     0     9     13
26         9         8        5     0     4     15
27         4         5        6     0     3     18
28         2         8        7     0     7      9
29         6         4        3     0     2     10
30         4         0        3     0     2      5
31         0        10        2     0     2     11
32         6         2        3     0     0      5
33         1         1        2     0     1      8
34         3         3        1     0     2      5
35         1         2        0     0     0      3
36         1         2        0     0     0      2
37         0         2        0     0     0      2
38         0         1        0     0     0      1
39         0         1        0     0     0      0
40         0         0        0     0     0      1
41         0         0        0     0     0      2
42         0         1        0     0     0      1
43         1         1        1     0     1      1
44         0         0        0     0     0      0
45         2         0        0     0     0      1
46         0         0        0     0     0      0
47         1         0        0     0     0      0
48         0         0        0     0     0      1
49         1         0        0     0     0      0
50         0         0        0     0     0      0
51         0         0        0     0     0      0
52         0         0        0     0     0      0
53         0         0        0     0     0      0
54         0         0        0     0     0      0
55         0         0        0     0     0      0
56         0         0        0     0     0      0
57         0         0        0     0     0      0
58         0         0        0     0     0      0
59         0         0        0     0     0      0
60         0         0        0     0     0      0
61         0         0        0     0     0      0
62         1         1        1     0     0      0
63         0         0        0     0     0      0
64         0         0        0     0     0      0
65         0         0        0     0     1      0
66         0         1        0     0     0      0
67         0         0        0     0     0      1
68         0         0        0     0     0      0
69         0         0        0     0     0      0
70         0         0        0     0     0      0
71         0         0        0     0     0      0
72         0         0        0     0     0      0
73         0         0        0     0     0      0
74         0         0        0     0     0      0
75         0         0        0     0     0      0
76         0         0        0     0     0      0
77         0         0        0     0     0      0
78         0         1        0     0     0      0
79         0         0        0     0     0      0
80         0         0        0     0     1      0
81         0         0        0     0     0      0
82         0         0        0     0     0      0
83         0         0        0     0     0      0
84         0         0        0     0     0      0
85         0         0        0     0     0      0
86         0         0        0     0     0      0
87         0         0        0     0     0      0
88         0         0        0     0     0      0
89         0         0        0     0     0      0
90         0         0        0     0     0      1
}\hallucinatedclaims

Table~\ref{tab:eval-limitations} gives examples of non-claims, repeated claims, and irrelevant claims that \veriscore{} counts as supported.

\subsection{Sensitivity to the Recall Normalizer}
\label{app:ksens}

Adding the \factgrpo{} rerun to the pool moves its median claim count to 23 for biographies and 21 for Alpaca-Fact. At those values every $\text{F1}@K$ falls by about .01 to .02, the order of the methods stays the same apart from one tie on Alpaca-Fact, and the one significant gain of \factgrpo{} no longer survives the paired test, so we keep $K$ fixed rather than let one run move every other row.

\begin{table}[H]
\centering
\caption{$\text{F1}@K$ of the same responses under the per-domain $K$ produced by different model pools scored in this study, given as biography / Alpaca-Fact. From left to right, the pools are a twelve-model pool with five \faith{} variants, a ten-model pool, a nine-model pool, the six-run pool of the earlier scoring pass, and a single model scored alone. The responses are those of the earlier scoring pass, with the first \factgrpo{} run. Precision does not depend on $K$.}
\label{tab:ksens}
\small
\setlength{\tabcolsep}{4pt}
\begin{tabular}{@{}l ccccc@{}}
\toprule
& \multicolumn{5}{c}{$K$ (biography / Alpaca-Fact)} \\
\cmidrule(lr){2-6}
Method & 5 / 7 & 14 / 15 & 20 / 18 & 23 / 21 & 29 / 24 \\
\midrule
\multicolumn{6}{@{}l}{Biography ($n = 492$)} \\
base        & .442 & .398 & .371 & .357 & .331 \\
\flame      & .435 & .394 & .367 & .353 & .326 \\
\factalign  & .427 & .384 & .357 & .342 & .315 \\
\knowrl     & .421 & .378 & .350 & .336 & .311 \\
\factgrpo   & .464 & .430 & .399 & .382 & .347 \\
\faith      & .164 & .068 & .049 & .043 & .034 \\
\addlinespace
\multicolumn{6}{@{}l}{Alpaca-Fact ($n = 249$)} \\
base        & .684 & .635 & .615 & .594 & .574 \\
\flame      & .681 & .629 & .609 & .588 & .568 \\
\factalign  & .679 & .624 & .604 & .584 & .564 \\
\knowrl     & .675 & .625 & .605 & .585 & .564 \\
\factgrpo   & .704 & .661 & .641 & .622 & .600 \\
\faith      & .048 & .026 & .022 & .019 & .017 \\
\bottomrule
\end{tabular}
\end{table}

\begin{table*}[t]
\centering
\small
\setlength{\tabcolsep}{4pt}
\renewcommand{\arraystretch}{1.15}
\begin{tabularx}{\textwidth}{@{}>{\raggedright\arraybackslash}p{0.155\textwidth}X>{\raggedright\arraybackslash}p{0.27\textwidth}@{}}
\toprule
\textbf{Question} & \textbf{Response claims} & \textbf{Comment} \\
\midrule
Tell me a bio of Mindy Smith.
& 
    [\factgrpo{}]
    \begin{itemize}[leftmargin=*, nosep]
        \item[\ding{55}] \textbf{Mindy Smith has released several singles throughout her career. Below is a list of some of her notable releases:}
        \item \textbf{- ``One Moment More''}
        \item \textbf{- ``It's That Time Again''}
        \item \textbf{- ``Come to Jesus''}
        \item \textbf{Mindy Smith has been performing at festivals and venues across the 
        United States.}
    \end{itemize}
    ...
&  \veriscore{} extracts bare song titles as check-worthy claims and labels them supported. \\
\addlinespace
Tell me a bio of Jessie Mae Brown Beavers.
&
    [\factalign{}]
    \begin{itemize}[leftmargin=*, nosep]
        \item[\ding{55}] \textbf{Jessie Mae Brown Beavers was active in the Civil Rights Movement in Mississippi.}
        \item \textbf{Jessie Mae Brown Beavers was a community leader.}
        \item \textbf{Jessie Mae Brown Beavers was an community leader.}
        \item \textbf{Jessie Mae Brown Beavers was an activist.}
    \end{itemize}
    ...
&  Near-duplicate claims are each counted as separately supported, inflating the factuality score. \\
\addlinespace
Tell me a bio of Diego Suarez (garden designer).
&
    [\factgrpo{}]
    \begin{itemize}[leftmargin=*, nosep]
        \item \emph{Garden designers design and plan gardens.}
        \item \emph{Garden designers consider the climate when designing gardens.}
        \item \emph{Garden designers consider the soil type when designing gardens.}
        \item \emph{Garden designers aim to achieve a desired aesthetic in their designs.}
    \end{itemize}
    ...
&  Generic filler claims are factually supported but irrelevant to the question. \factgrpo{} exploits such claims to hack the reward (Table~\ref{tab:rfact-dominates-rrel}). \\
\bottomrule
\end{tabularx}
\caption{Examples of limitations in claim-level factuality evaluation. Non-claims, repeated claims, and irrelevant claims are each counted as supported, inflating the factuality score. A cross (\ding{55}) in place of the bullet marks claims that \veriscore{} labels not supported (all other claims are labeled supported), and typeface marks the relevance judgment (\textbf{bold}: relevant, \emph{italic}: irrelevant).}
\label{tab:eval-limitations}
\end{table*}

\subsection{Reference Verification Against Wikipedia}
\label{app:reference}

We re-verify biography claims against the subject's article in a 2023 English Wikipedia dump, using BM25 over 25-word windows and the same verifier (Table~\ref{tab:reference_bio}). Precision falls by about twenty points for methods producing long responses. Wikipedia supports only two fifths of web-supported claims, while nearly all Wikipedia-supported claims also have web support. \factgrpo{} and \flame{} move from below to above the base model, significantly so for \factgrpo{}. \knowrl{} remains significantly below it, while \faith{} retains the highest precision with very few claims.

\begin{table}[H]
\centering
\small
\caption{Comparison between evidence sources for verification. Biography claims of each method verified against web search vs against the subject's Wikipedia article with the same verifier, on the 489 prompts with an article that every method answered. \#Sup.\ is the mean number of supported claims per response, P the mean per-response precision, and Agreement the share of claims with the same verdict under both sources. † marks a significant difference from the base row (paired bootstrap, Holm-corrected $p < 0.05$).}
\label{tab:reference_bio}
\footnotesize
\setlength{\tabcolsep}{3pt}
\begin{tabular}{@{}lrrrrrr@{}}
\toprule
 & & \multicolumn{2}{c}{Web search} & \multicolumn{2}{c}{Wikipedia} & \\
\cmidrule(lr){3-4} \cmidrule(lr){5-6}
Method & \#Claims & \#Sup. & P & \#Sup. & P & Agree. \\
\midrule
\multicolumn{7}{l}{\textit{Biography ($n = 489$)}} \\
base (untrained) & 27.6 & 12.0 & .351 & 5.5 & .140 & .729 \\
\midrule
\flame           & 23.6 & 10.6 & .338 & 5.1 & .149 & .729 \\
\factalign       & 27.1 & 11.7 & .344 & 5.5 & .141 & .734 \\
\knowrl          & 29.8 & 11.3 & .281\rlap{$^{\dagger}$} & 5.1 & .116\rlap{$^{\dagger}$} & .760 \\
\factgrpo        & 25.4 & 10.1 & .342 & 4.7 & .155\rlap{$^{\dagger}$} & .752 \\
\faith           & 0.6 & 0.4 & .375 & 0.3 & .316\rlap{$^{\dagger}$} & .841 \\
\bottomrule
\end{tabular}

\end{table}

\subsection{Human Audit of Claim Labels}
\label{app:audit}

The audit sample has 300 claims, 50 from each of the six variants of the common benchmark, with 200 from biography prompts and 100 from Alpaca-Fact prompts.
Within each variant and dataset, the claims are spread evenly over the four combinations of the \veriscore{} verdict and the relevance label, and at most one claim is drawn from a response.
Since the combinations are balanced, the agreement on the sample is not the agreement on all claims.
Two of the authors labeled the sample independently. For each claim, each annotator sees the prompt, the claim, the sentence it was extracted from with its neighbors, up to ten web search results, and, for biographies, up to ten passages from the subject's Wikipedia article.
The automatic labels are not shown.

Each annotator gives each claim up to three labels.
The first two judge whether the web evidence and the Wikipedia evidence support the claim, each with the labels \emph{Supported}, \emph{Not supported}, and \emph{Cannot tell from evidence}.
A claim is supported when the evidence states or clearly implies every part of it, including names, dates, numbers, and places, and it is not supported when the evidence contradicts any part.
When the evidence neither supports nor contradicts the claim, for example because it is about something else, too vague, or cut off, the label is \emph{Cannot tell from evidence}, and the same holds when only part of the claim is covered or when the evidence conflicts.
Annotators use only the evidence in the column being judged, not their own knowledge or a search, and treats evidence about a different entity with the same name as neither supporting nor contradicting the claim.

The third label judges whether the claim is \emph{Relevant} or \emph{Irrelevant} to the prompt, treating the claim as true.
A claim is relevant when it gives information that the prompt asks for about its subject, such as any fact about the life, family, education, career, work, or legacy of the person in a biography prompt.
It is irrelevant when it is mainly about something else, such as a related person, place, organization, or event, when it falls outside the scope of the prompt or adds detail that was not asked for, or when it is a statement about the answer, the task, or the model.
When unsure, annotators choose \emph{Irrelevant}, and a free-text note flags claims that are garbled or are not claims at all.
After a check of the first annotator's labels against these rules flagged blank labels and labels that contradict a rule, that annotator revised the labels they agreed with.

Table~\ref{tab:audit} gives the agreement of each annotator with the automatic labels, of the two annotators with each other, and of their consensus, the claims to which both gave the same label, with the automatic labels. On the consensus claims, the share of each automatic label that the annotators confirm is .97 for supported and .65 for unsupported under \veriscore{}, .99 for relevant and .55 for irrelevant under the LLM labels, and .76 for supported and .78 for unsupported under the Wikipedia verdicts.

\begin{table}[H]
\centering
\caption{Agreement of the human audit with the automatic labels. Agr.\ is the share of claims with the same label and $\kappa$ Cohen's kappa. Support rows count \emph{Cannot tell from evidence} as unsupported, as the verifiers do.}
\label{tab:audit}
\small
\setlength{\tabcolsep}{4pt}
\begin{tabular}{@{}l rrr@{}}
\toprule
& $n$ & Agr. & $\kappa$ \\
\midrule
\multicolumn{4}{@{}l}{Web support vs.\ \veriscore{}} \\
\quad first annotator & 261 & .797 & .58 \\
\quad second annotator & 289 & .751 & .49 \\
\quad between annotators & 261 & .778 & .50 \\
\quad consensus & 203 & .842 & .65 \\
\addlinespace
\multicolumn{4}{@{}l}{Relevance vs.\ LLM labels} \\
\quad first annotator & 267 & .835 & .62 \\
\quad second annotator & 296 & .780 & .53 \\
\quad between annotators & 266 & .842 & .54 \\
\quad consensus & 224 & .862 & .62 \\
\addlinespace
\multicolumn{4}{@{}l}{Wikipedia support vs.\ reference verdict} \\
\quad first annotator & 173 & .792 & .50 \\
\quad second annotator & 192 & .677 & .32 \\
\quad between annotators & 173 & .821 & .64 \\
\quad consensus & 142 & .775 & .50 \\
\bottomrule
\end{tabular}

\end{table}

\section{Reproducibility Details}
\label{app:repro}

The repository's reproducibility document records settings from saved arguments, scripts, logs, and outputs; unrecorded values are marked missing. It lists model revisions, software versions, repository commits, hardware, training times, seeds, hyperparameters, decoding settings, dataset selection, scoring configurations, and API evaluation dates.

In the common evaluation, \flame{} reuses the 70B reproduction's SFT data with Llama-3.1 prompts. Its targets are therefore Llama-2-70B generations and human responses, not the current policy's generations. It also trains for 1,000 rather than 500 steps.

Training used about 140 GPU hours for the common evaluation and 627 for the original setups, including 547 for \flame{} at 70B. Evaluation used at least 70 GPU hours. API costs are estimated from call counts and list prices because billed amounts were not recorded.

Three details qualify Section~\ref{sec:setup}. State-prompted \faith{} uses limits of 512 long-form and 64 short-form tokens; neither is reached. The \flame{} classifier selects 253 of 805 AlpacaEval instructions, against 241 in \citet{Lin:2024:NeurIPS}. Long-form tables exclude ten of 748 \factgrpo{} rerun responses that end within the reasoning trace without an answer.

\subsection{Data, Metrics, and Training Stages}

Table~\ref{tab:suite} lists the base models, evaluation datasets, and factuality metrics of the five papers, and Table~\ref{tab:pipelines} their training stages.

\begin{table*}[t]
\centering
\caption{Base models, datasets, and metrics of the five methods. None shares a model or dataset with all others. Three metrics are precision-like, one credits refusals, and one excludes them.}
\label{tab:suite}
\small
\setlength{\tabcolsep}{3pt}
\begin{tabular}{@{}l >{\raggedright\arraybackslash}p{0.14\textwidth} >{\raggedright\arraybackslash}p{0.30\textwidth} >{\raggedright\arraybackslash}p{0.21\textwidth}@{}}
\toprule
\textbf{Method} & \textbf{Base models} & \textbf{Evaluation data} &
\textbf{Factuality metric} \\
\midrule
\flame~\citep{Lin:2024:NeurIPS}
  & Llama-2 7B, 70B
  & Bio~\citep{Min:2023:EMNLP}, Alpaca-Fact~\citep{Lin:2024:NeurIPS}, FAVA~\citep{Mishra:2024:arXiv}
  & FActScore (Inst-LLAMA${+}$npm) \\
  & & TruthfulQA~\citep{Lin:2022:ACL}, AlpacaEval~\citep{Dubois:2024:arXiv} & BLEU/ROUGE; GPT-4 win rate \\
\addlinespace
\factalign~\citep{Huang:2024:EMNLP}
  & gemma-2b, \textbf{Llama-3-8B}, Phi3-Mini
  & LongFact-Objects~\citep{Wei:2024:NeurIPS}
  & SAFE F1@100 (ColBERT) \\
  & & FActScore biographies~\citep{Min:2023:EMNLP} & FActScore (GPT-3.5 assessor) \\
\addlinespace
\knowrl~\citep{Ren:2026:ACL}
  & R1-Distill-Qwen-7B, Skywork-OR1-7B
  & SimpleQA~\citep{Wei:2024:arXiv}, ChineseSimpleQA~\citep{He:2024:arXiv}
  & accuracy, non-rejected only \\
  & & TruthfulQA~\citep{Lin:2022:ACL}, GPQA~\citep{Rein:2023:arXiv}, AIME & BLEU; accuracy \\
\addlinespace
\factgrpo~\citep{Chen:2026:ICML}
  & Llama-3.1-8B-Instruct
  & LongFact~\citep{Wei:2024:NeurIPS}, FAVA~\citep{Mishra:2024:arXiv}, Alpaca-Fact~\citep{Lin:2024:NeurIPS}, Bio~\citep{Min:2023:EMNLP}, FactBench~\citep{Bayat:2024:arXiv}, Factory~\citep{Chen:2025:arXiv}
  & \veriscore{} prec., F1@K \\
\addlinespace
\faith~\citep{Dong:2026:ACL}
  & \textbf{Llama-3-8B}, Mistral-7B
  & TriviaQA~\citep{Joshi:2017:ACL}, SciQ~\citep{Welbl:2017:WNUT}, NQ-Open~\citep{Kwiatkowski:2019:TACL}
  & precision over \emph{known} \\
  & & WebQuestions~\citep{Berant:2013:EMNLP} (OOD) & truthfulness, \emph{credits refusals} \\
\bottomrule
\end{tabular}
\end{table*}

\begin{table*}[t]
\centering
\caption{Training stages and optimization signals. Implicit rewards use $\beta\log(\pi/\pi_{\text{ref}})$; rule-based rewards use verifier outputs and checks; learned rewards use a trained model.}
\label{tab:pipelines}
\small
\setlength{\tabcolsep}{4pt}
\begin{tabular}{@{}l p{0.30\textwidth} l p{0.30\textwidth} l@{}}
\toprule
\textbf{Method} & \textbf{Training stages} & \textbf{Signal} &
\textbf{What it optimizes} \\

\midrule
\flame
  & factuality-aware SFT $\to$ factuality-aware DPO
  & implicit
  & preference pairs scored with \textsc{FActScore} against retrieval
   \\
\addlinespace
\factalign
  & fKTO, from an instruction-tuned checkpoint that supplies the
    supervised stage
  & implicit
  & $\mathcal{L}_{\text{KTO}} + \lambda\,\mathcal{L}_{\text{fKTO}}$:
    response-level KTO plus sentence-level labels
  \\
\addlinespace
\knowrl
  & cold-start SFT $\to$ GRPO
  & rule
  & format $\pm 1$, plus an LLM verdict $({+}2/{-}1/{+}1)$, plus
    \textsc{FActScore} partial credit when wrong and not a refusal
   \\
\addlinespace
\factgrpo
  & SFT (3k) $\to$ Dr.\ GRPO
  & rule
  & $-1$ if malformed, else $F/(T{+}1) + \lambda\log(1{+}F) + \mu w$,
    with $\lambda_{\text{len}}{=}0$, $\lambda_{\text{llm}}{=}0.1$; $S$, $T$ from \veriscore
   \\
\addlinespace
\faith
  & estimator $\to$ SFT $\to$ reward model $\to$ PPO $\to$ RAG
    rectifier (RAFT)
  & learned
  & a Llama-3-8B reward model with a scalar value head, trained to
    predict the \texttt{EM\_RP} containment score
   \\
\bottomrule
\end{tabular}
\end{table*}

\end{document}